\documentclass[10pt,twocolumn,letterpaper]{article}

\usepackage{cvpr}              % To produce the CAMERA-READY version
\usepackage{fancyhdr}
\usepackage{graphicx}
\usepackage{amsthm}
\usepackage{algorithm}
\usepackage{algorithmic}
\usepackage{multirow}
\usepackage{makecell}
\usepackage{graphicx}
\usepackage{subcaption}
\usepackage{xcolor}
\usepackage{pifont}
\newtheorem{theorem}{Theorem}[section]

\theoremstyle{definition}           % 用直立体
\newtheorem{definition}[theorem]{Definition}

\theoremstyle{remark}

\definecolor{cvprblue}{rgb}{0.21,0.49,0.74}
\usepackage[pagebackref,breaklinks,colorlinks,allcolors=cvprblue]{hyperref}

\def\paperID{4704} % *** Enter the Paper ID here
\def\confName{CVPR}
\def\confYear{2026}
\title{COPO: Causal-Oriented Policy Optimization for Hallucinations of MLLMs \thanks{Preprint. Under Review.}}

\author{\textbf{Peizheng Guo}\textsuperscript{\rm 1,\rm 2}\thanks{These authors contributed equally.}, 
    \textbf{Jingyao Wang}\textsuperscript{\rm 1,\rm 2}\footnotemark[2], 
    \textbf{Wenwen Qiang}\textsuperscript{\rm 1,\rm 2}\thanks{Corresponding author.}, 
    \textbf{Jiahuan Zhou}\textsuperscript{\rm 3}, \\
    \textbf{Changwen Zheng}\textsuperscript{\rm 1,\rm 2}, 
    \textbf{Gang Hua}\textsuperscript{\rm 4}\and
    \textsuperscript{\rm 1}University of Chinese Academy of Sciences,
    \textsuperscript{\rm 2}Institute of Software Chinese Academy of Sciences,\\
    \textsuperscript{\rm 3}Wangxuan Institute of Computer Technology, Peking University,\\
    \textsuperscript{\rm 4}Amazon.com, Inc., Bellevue, WA, 98004, USA\\
{\tt\small guopeizheng2025@iscas.ac.cn}
}

\begin{document}
\maketitle
\begin{abstract}
Despite Multimodal Large Language Models (MLLMs) having shown impressive capabilities, they may suffer from hallucinations. 
Empirically, we find that MLLMs attend disproportionately to task-irrelevant background regions compared with text-only LLMs, implying spurious background-answer correlations. 
We claim and analyze that (i) outcome-based rewards can be an important factor leading to spurious correlations, and (ii) spurious correlations can be an important factor leading to hallucinations. 
Based on these results, we propose Causal-Oriented Policy Optimization (COPO) to mitigate these spurious correlations, thus addressing the issue of hallucinations. It imposes token-level sufficiency and necessity constraints to measure each inference token's causal contribution, thus ensuring correct and evidence-grounded output.
Specifically, we first evaluate each token's causal contribution via a newly proposed causal completeness reward. This reward is then used to construct a causally informed advantage function within the GRPO optimization framework, encouraging the model to focus on tokens that are causally sufficient and necessary for accurate generation. Experimental results across various benchmarks demonstrate the advantages of COPO.
\end{abstract}
\section{Introduction}
\label{sec:introduction}
\begin{figure}
    \centering
    \includegraphics[width=\columnwidth]{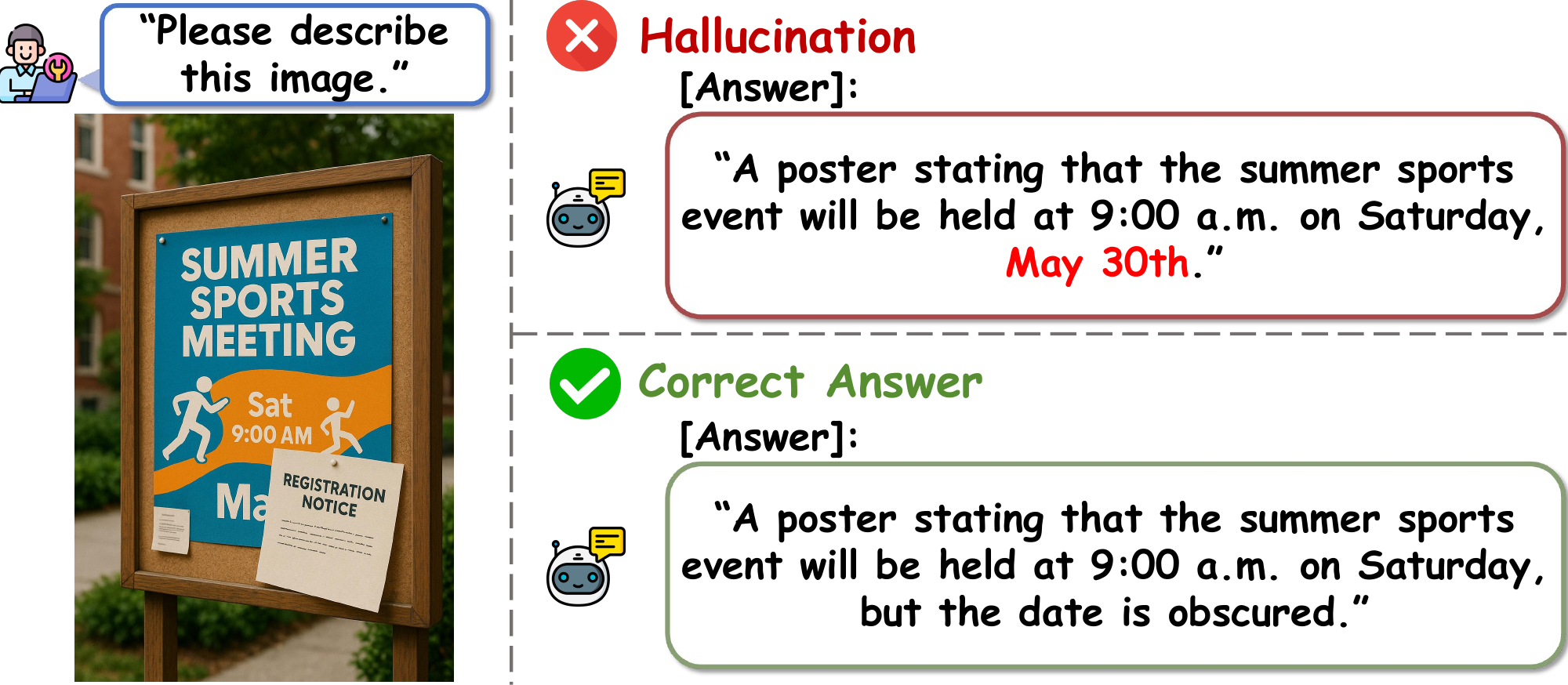}
    \caption{
    Example of hallucination in MLLMs: invent a specific date when the event date in the poster is actually obscured.
    }
    \label{fig:cat_sofa}
\end{figure}

In the context of Multimodal Large Language Models (MLLMs), hallucination refers to the generation of content that is inconsistent with, or unsupported by, the available input evidence across modalities (e.g., images and text) and any permissible external knowledge sources \cite{bai2024hallucination,huang2025survey}. Specifically, a hallucination occurs when the model produces descriptions that either contradict the observable information in the input or present unverifiable details with unwarranted confidence \cite{li2023evaluating,zhou2023analyzing,bai2024hallucination}. This phenomenon reflects a failure of cross-modal grounding, i.e., the model relies on spurious correlation or prior biases instead of faithfully integrating the evidence provided \cite{li2023evaluating,zhou2023analyzing,bai2024hallucination}. By contrast, outputs that explicitly acknowledge uncertainty (e.g., ``the date is unclear'' or ``the object is partially occluded'') are not considered hallucinations \cite{kalai2025languagemodelshallucinate}, as they demonstrate calibrated reasoning aligned with the evidence. 
As shown in \textbf{Figure~\ref{fig:cat_sofa}}, in an image captioning task, when presented with an image of a summer sports event poster where the specific date is obscured, an ideal output may be ``a poster stating that the summer sports event will be held at 9:00 a.m. on Saturday, but the date is obscured''. However, it may generate a hallucinated description like ``...the summer sports event will be held at 9:00 a.m. on Saturday, May 30th''. 

\begin{figure*}[t]
    \centering
    \begin{subfigure}[t]{0.49\textwidth}
        \centering
        \includegraphics[width=\textwidth]{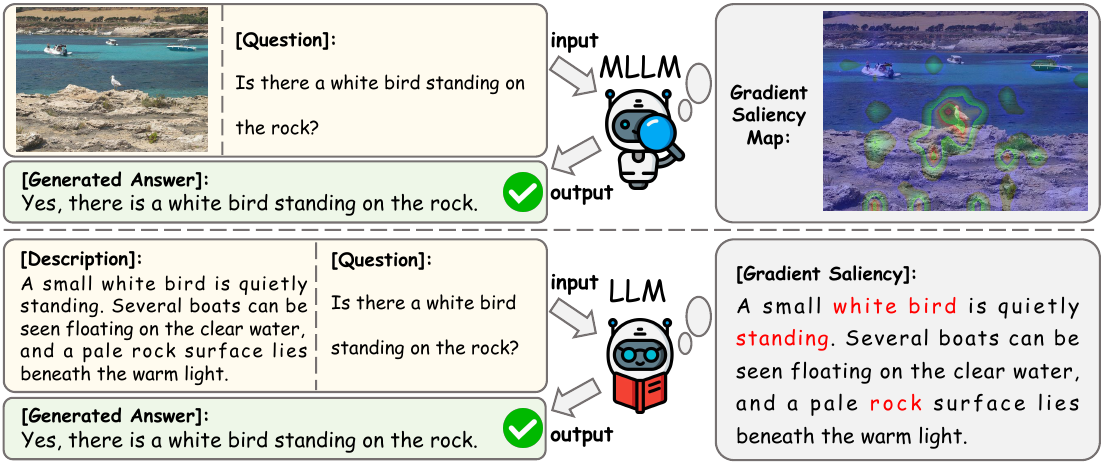}
        \caption{Scenario 1}
        \label{fig:motivation_1}
    \end{subfigure}
    \hfill
    \begin{subfigure}[t]{0.49\textwidth}
        \centering
        \includegraphics[width=\textwidth]{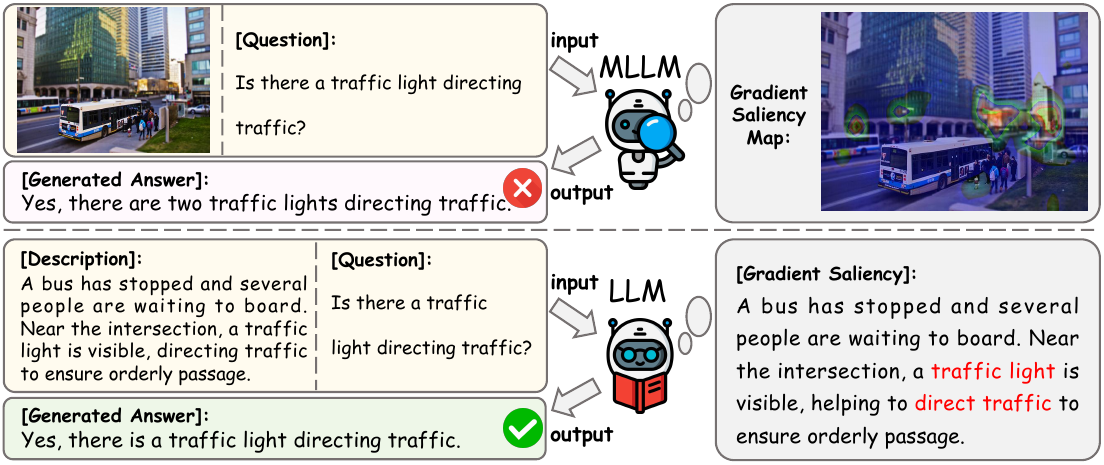}
        \caption{Scenario 2}
        \label{fig:motivation_2}
    \end{subfigure}
    \caption{Motivating results. Both MLLM and LLM are trained via GRPO. }
    \label{fig:motivation}
\end{figure*}

An effective way to alleviate hallucinations is to enhance the reasoning ability of MLLMs \cite{ahmadian2024back,lambert2024tulu,shao2024deepseekmath}. 
Reinforcement learning (RL) has emerged as a mainstream post-training paradigm for this purpose. It optimizes models based on reward signals derived from verifiable outcomes (e.g., final-answer correctness), enabling a scalable mechanism for eliciting effective problem-solving behaviors \cite{guo2025deepseek,xie2025logic,deng2025supervisedreinforcementlearningexpert}. Among various RL methods, Group Relative Policy Optimization (GRPO) is widely adopted for its effectiveness \cite{dai2025qoq,wu2025spatial}. It encourages the model to explore diverse reasoning paths and iteratively refine them through the outcome reward that evaluates each path's quality \cite{yang2025ui,park2025vidguard}.

Generally, GRPO is trained using outcome-based rewards and was originally proposed for optimizing Large Language Models (LLMs) \cite{guo2025deepseek}. However, unlike LLMs that rely solely on the language modality, MLLMs additionally process visual inputs that carry much richer and denser information. For instance, while describing a dog in text may require several sentences, a single image can immediately convey its appearance, posture, and surroundings in detail.
Through our experiments, we find that when such information-rich visual input is introduced, an MLLM trained only with outcome-based rewards under the GRPO framework tends to exploit spurious correlations during inference. Specifically, we trained an MLLM (with both image and text inputs) and an LLM (with text only) under the same GRPO setting and analyzed their outputs and gradient patterns. As shown in \textbf{Figure \ref{fig:motivation}}, we observe that regardless of whether the answer is correct or incorrect, the gradient saliency over background regions in the MLLM is consistently higher than that in the LLM.
This empirical result suggests a spurious dependency on task-irrelevant background information.

In this paper, we argue that: (i) outcome-based rewards can be an important factor leading to spurious correlations; and (ii) spurious correlations may in turn lead to hallucinations. We first explain the first argument. During GRPO training, rewards are assigned only to the final outcome, so the MLLM is reinforced whenever the final answer is correct, even if the reasoning process partially depends on background elements in the visual input. Over time, this can gradually encourage the model to associate irrelevant background cues with correct outcomes. When the input consists of a limited number of visual samples, a large amount of background information may overlap with the foreground cues, making it difficult for the MLLM to avoid considering the background information during the decision-making process. Next, we explain the second argument. One direct piece of evidence comes from prior studies \cite{rohrbach2018object,leng2024mitigating,kaul2024throne,wysoczanska2025ovfact,kalai2025languagemodelshallucinate}, which have shown that such spurious correlations can lead to outputs that contradict factual reality. We can further analyze this with a finer level of detail. Specifically, in the inference stage, token generation usually follows sampling strategies such as Top $K$ \cite{holtzman2019curious}, and the final answer is selected from multiple candidate responses using approaches like beam search \cite{vinyals2015show}, where each candidate consists of a sequence of tokens generated by these probabilistic procedures. Given this generation process and the fact that the model simultaneously attends to both foreground and background information, there is a non-negligible chance that background features become dominant during candidate generation or selection. When this occurs, the model may produce outputs that deviate from factual reality, thus manifesting as hallucinations.

\begin{figure}[t]
    \centering
    \begin{subfigure}[t]{0.22\textwidth}
        \centering
        \includegraphics[width=0.9\textwidth]{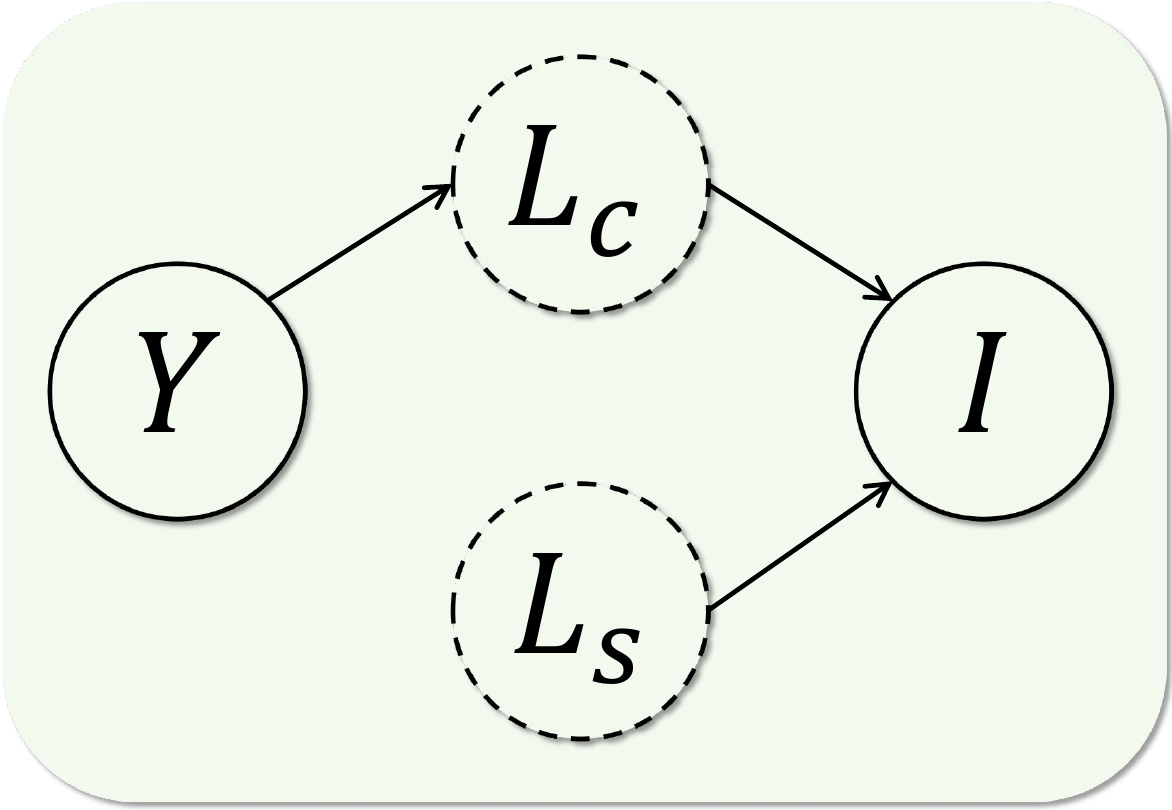}
        \caption{Data generation process}
        \label{fig:scm_mml_data}
    \end{subfigure}
    \hfill
    \begin{subfigure}[t]{0.22\textwidth}
        \centering
        \includegraphics[width=0.9\textwidth]{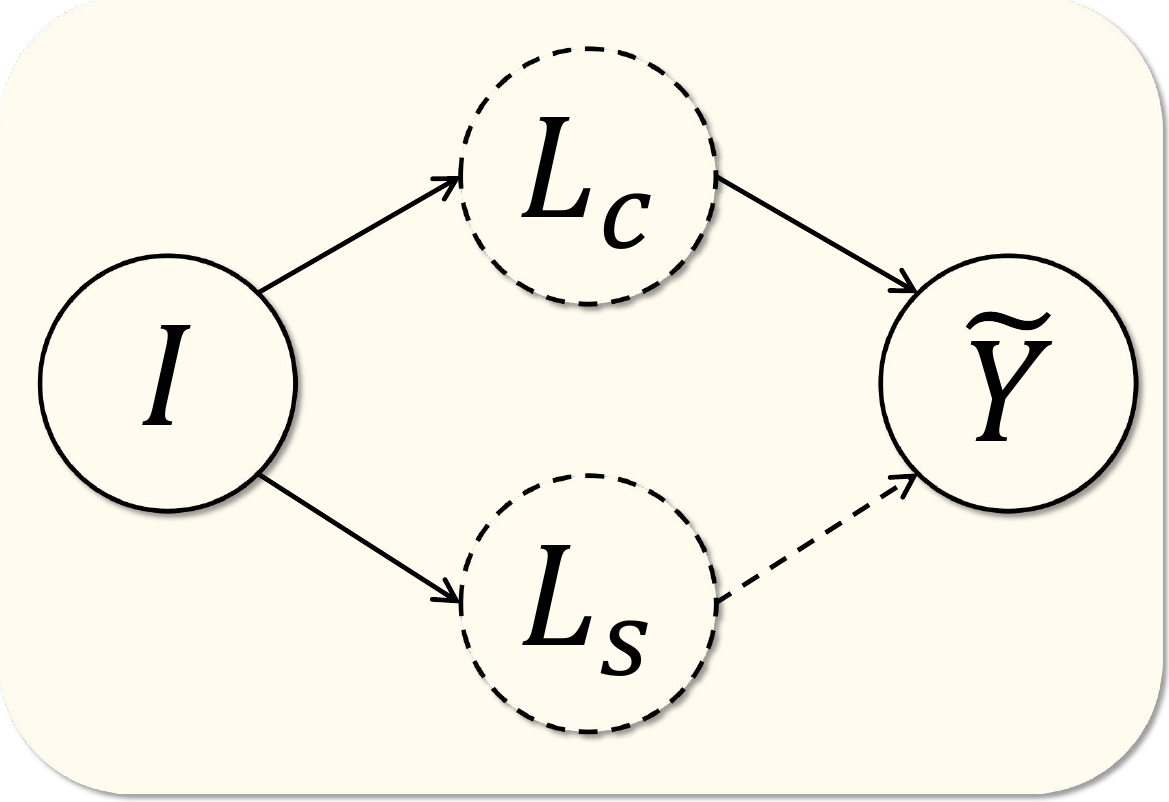}
        \caption{Token generation of MLLMs}
        \label{fig:scm_mllm_inference}
    \end{subfigure}
    \caption{SCMs for MLLMs. The solid circles denote observable variables, dashed circles as unobservable variables, black arrows as true correlations, and dashed arrows as spurious correlations.}
    \label{fig:scm_mllm}
\end{figure}

Building on the above analysis, we first build Structural Causal Models (SCMs) for MLLMs. As shown in \textbf{Figure \ref{fig:scm_mllm}}, let $Y$ be the ground-truth answer and $I=(I_v,I_t)$ the image–text input. Following \cite{hu2022causal}, each sample is driven by two sets of latent factors: $L_c$, which are causally related to $Y$, and $L_s$, which are non-causal. 
\textbf{Figure \ref{fig:scm_mllm_inference}} illustrates the inference process: given $I$, the model produces a token sequence $o=\{o_1,\dots,o_T\}$, containing the predicted answer $\tilde{Y}$ and reasoning tokens $\bar{o}=\{o\}\setminus \{\tilde{Y}\}$. 
Under this SCM, $\tilde{Y}$ is expected to depend on $L_c$ and remain invariant to $L_s$. To achieve this, we introduce causal sufficiency and necessity constraints into the reward design. Causal sufficiency requires that keeping a reasoning token \( \bar{o}_t \) should increase the probability of obtaining the correct answer. Causal necessity, in contrast, requires that perturbing or removing \( \bar{o}_t \) should decrease or alter the correctness of the answer. 
By viewing the token sequence \( \bar{o} \) as a series of intermediate reasoning tokens, these constraints help quantify each token's causal contribution and guide the reward to favor evidence-grounded reasoning rather than background-dependent correlations. To implement this, we propose a causal-oriented policy optimization framework (COPO) for reducing hallucinations in MLLMs. 
The core of COPO is a causal completeness reward that measures whether each token is both sufficient and necessary for producing the correct output. Specifically, we first compute two token-level scores: a sufficiency score reflecting how much retaining a token improves answer correctness, and a necessity score reflecting how much perturbing or removing it degrades the result. These scores are then combined through a weighted formulation into the causal completeness reward, so that tokens that are both independently helpful and indispensable receive higher rewards. We integrate this reward into GRPO by adjusting the advantage calculation, steering policy learning toward tokens that contribute causally to the final answer and away from spurious background cues. Extensive experiments demonstrate the advantages of COPO.

The main contributions can be summarized as: (i) Through causal analysis and empirical evidence, we reveal that the outcome-only rewards of GRPO may reinforce spurious correlation on non-causal background signals in MLLMs, producing seemingly correct but unsupported hallucinated outputs. (ii) We propose COPO, a causal-oriented policy optimization framework that embeds token-level sufficiency and necessity into a composite causal-completeness reward. By integrating it into the advantage of GRPO, we promote evidence-grounded tokens and reduce hallucinations. (iii) Extensive experiments conducted on various benchmark datasets demonstrate the advantages of the proposed COPO.

\section{Related Work}
\label{sec:related_work}

\paragraph{Hallucinations of MLLMs}
\label{sec:related_work_mllm}
MLLMs jointly process visual and textual inputs, achieving great performance on various tasks \cite{li2023blip,wang2024meta,liu2023visual,zhu2023minigpt,wang2025learning,dai2023instructblipgeneralpurposevisionlanguagemodels,wang2023amsa,chen2023shikra}. Despite strong task performance, MLLMs remain prone to hallucinations.
A hallucination is model-generated content that contradicts or is unsupported by available input evidence \cite{bai2024hallucination,huang2025survey}.
Recent works \cite{chen2025perturbollava,yu2024hallucidoctor,liu2023mitigating,leng2024mitigating,qu2024alleviating} propose to optimize the reasoning capability of MLLMs to reduce hallucinations. GRPO, a post-training method, is widely adopted for its effectiveness \cite{dai2025qoq,wu2025spatial}. 
While GRPO optimizes reasoning via outcome-based rewards \cite{guo2025deepseek}, our experiments show that applying GRPO to MLLMs may induce spurious reliance on task-irrelevant visual cues, a behavior correlated with hallucinated outputs.
Consequently, in this paper, we explore from a causal view and propose a causal-oriented policy optimization framework to reduce hallucinations. More discussion and comparisons are shown in the \textbf{Appendix \ref{sec_app:discussion}}.

\vspace{-0.1in}

\paragraph{Causal Theory}
\label{sec:related_work_causality_theory}
Causal theory provides a principled framework for reasoning, forming the theoretical foundation for addressing complex decision-making problems beyond mere correlation-based analysis \cite{pearl2009causality,neuberg2003causality,pearl2009causality,wang2024hacking}. 
Some works explore the integration of causality into LLMs, e.g., \cite{kiciman2023causal} shows that while LLMs are not inherently causal models, they can be adapted to perform causal tasks by aligning representations with causal quantities; \cite{gu2025group} incorporates a structural causal model into policy optimization, addressing dependencies among candidate responses via causal projection.
Different from these works, we focus on the hallucination of MLLMs, analyzing from both empirical and causal perspectives. We propose the causal sufficiency and necessity constraints into a novel causal-orient policy optimization framework to mitigating hallucinations. See the \textbf{Appendix} for more analyses and comparisons.

\section{Problem Setting and Analysis}
\label{sec:problem_setting_analysis}
In this section, we first introduce the problem settings of MLLMs in \textbf{Subsection \ref{sec:problem_setting}}. 
Next, we provide the empirical and causal analyses regarding the hallucinations of MLLMs in \textbf{Subsection \ref{sec:empirical}} and \textbf{Subsection \ref{sec:cc}}, respectively.

\subsection{Problem Setting}
\label{sec:problem_setting}
Given a dataset $\mathcal{D}=\{(I_v^{(n)},I_t^{(n)},Y^{(n)})\}_{n=1}^N$ consists of a set of paired image-text samples, where each pair $I^{(n)}=(I_v^{(n)},I_t^{(n)})$ consists of a image $I_v^{(n)}$ and a text prompt $I_t^{(n)}$, and $Y^{(n)}$ is the ground-truth answer. 
For each input $I^{(n)}=(I_v^{(n)},I_t^{(n)}), \;n\in \{1,\cdots,N\}$, GRPO samples a set of candidate outputs $\{{o^i\}}_{i=1}^G$ from the old policy $\pi_{\theta_{\rm old}}(\cdot|I^{(n)})$. For a sequence $o^i=\{o^i_1,\cdots,o^i_{T_i}\}$, it contains both intermediate reasoning steps and answer tokens. We denote the answer tokens as $\tilde{Y}^i=\{\tilde{y}^i_1,\cdots,\tilde{y}^i_{T_{y^i}}\}$, with the intermediate reasoning steps are $\{o^i\}\setminus \{\tilde{Y}^i\}$. 
The updated policy $\pi_\theta$ is obtained by maximizing the objective:
\begin{equation}\label{eq:grpo_mllm}
\begin{aligned}
\mathcal{J}_{\rm GRPO}
= \mathbb{E}&_{[I\sim P,\ \{o^i\}\sim \pi_{\theta_{\rm old}}]} \\
\frac{1}{G}\sum_{i=1}^{G}\frac{1}{T_i}\sum_{t=1}^{T_i}
 \min(R_{i,t}(\theta)&A_i,\ \Psi(A_{i,t}))
-\beta\; \mu(\pi_\theta)),\\
\end{aligned}
\end{equation}
where $\epsilon$ and $\beta$ are hyperparameters, $\Psi(A_{i,t})={\rm clip}(R_{i,t},1-\epsilon,1+\epsilon)\cdot A_{i,t}$, $\mu(\pi_\theta)=D_{\rm KL}(\pi_\theta\Vert \pi_{\rm ref})$, and $T_i$ is the length of $o^i=\{o^i_1,\cdots o^i_{T_i}\}$.
The relative advantage $A_i$ is calculated within each group to capture the comparative quality of outputs:
\begin{equation}\label{eq:adv}
A_i=\frac{r^i-\mathrm{mean}(r^1,\ldots,r^{G})}{\mathrm{std}(r^1,\ldots,r^{G})},
\end{equation}
where $r^i = \mathrm{reward}(Y^i)$ is the outcome reward \cite{guo2025deepseek}, which is calculated based on the correctness of $\tilde{Y}$ (e.g., $r^i=1$ when $\tilde{Y}$ is correct; otherwise, $r^i=0$), while $\mathrm{mean}(\cdot)$ and $\mathrm{std}(\cdot)$ are the mean and standard deviation over the reward group.
The token-level importance ratio is 
\begin{equation}\label{eq:ratio}
R_{i,j}(\theta)=\frac{\pi_\theta(o^{i}_t \mid I^{(n)},\ o^{i}_{<t})}{\pi_{\theta_{\rm old}}(o^{i}_t \mid I^{(n)},\ o^{i}_{<t})}.
\end{equation}
The KL term is computed token-wise with a fixed reference policy $\pi_{\rm ref}$ (typically $\pi_{\theta_{\rm old}}$) as
\begin{equation}\label{eq:kl}
\scalebox{0.93}{$D_{\rm KL}(\pi_\theta\Vert \pi_{\rm ref})=\frac{\pi_{\rm ref}(o^{i}_t | I^{(n)},\ o^{i}_{<t}\big)}{\pi_\theta(o^{i}_t | I^{(n)},\ o^{i}_{<t})}-\log\frac{\pi_{\rm ref}(o^{i}_t | I^{(n)},\ o^{i}_{<t})}{\pi_\theta(o^{i}_t | I^{(n)},\ o^{i}_{<t})}-1.$}
\end{equation}

\subsection{Empirical Findings and Analysis}
\label{sec:empirical}
Despite the strength of MLLMs, they may still face the challenge of hallucinations, i.e., generating content that deviates from factual reality \cite{li2023evaluating,liu2024survey,zhou2023analyzing}. 
In this section, we experimentally find an interesting phenomenon: under the post-training of GRPO, the gradient saliency of MLLMs on task-irrelevant background is higher than that of LLMs. Based on the empirical findings, we further present a possible explanation and analysis for hallucinations of MLLMs.

Specifically, we adopt GRPO post-training for both MLLMs (i.e., LLaVA \cite{liu2023visual}) and LLMs (i.e., LLaMA  \cite{grattafiori2024llama3herdmodels}) to ensure a consistent optimization scheme. 
For each example, the MLLM is provided with an image-text pair (e.g., an image with ``Is there a traffic signal light directing traffic?''), whereas the LLM is provided with the same question, accompanied by a textual description of the image. 
We require both models to generate short factual answers, and visualize their gradient distributions throughout the training process. By computing the average gradient across all training iterations, we generate gradient saliency maps for the MLLM and token embedding gradients for the LLM (with details in \textbf{Appendix \ref{sec_app:motivation_detail}}). 
From \textbf{Figure \ref{fig:motivation}}, we can observe that after the GRPO-based post-training, the MLLM exhibits high gradient saliency over both task-relevant regions and irrelevant background for answer generation, while the LLM's gradients concentrate on semantic elements directly tied to the question (e.g., ``white bird'', ``traffic light'').

\noindent\textbf{Empirical Analyses.} 
From the above observations, we speculate that this comparable gradient saliency between background and foreground cues in MLLMs may be a factor contributing to hallucinations in MLLMs. We further draw two claims: (i) the outcome-based rewards of GRPO may induce spurious correlations; and (ii) spurious correlations may produce hallucinations in MLLMs. 
Specifically, outcome-based rewards of GRPO are issued solely on final-answer correctness \cite{guo2025deepseek,wang2025learning,liu2025understanding}. Any generation trajectory that happens to yield a correct outcome receives positive reinforcement, even if it partially relies on task-irrelevant background cues. 
Our empirical results show that MLLMs may often exploit task-irrelevant background cues. For example, gradient saliency maps often show similar emphasis on background and foreground regions.
Then, over repeated updates via GRPO, the outcome-only feedback may promote the model to treat irrelevant background cues as predictive, strengthening spurious correlations between background signals and correct answers. In short, outcome-based rewards can create spurious correlations that entangle irrelevant background components with successful predictions. When given a limited number of visual samples as input, a significant amount of background information often overlaps with the foreground cues. This overlap makes it challenging for the MLLM to completely exclude the influence of background information. Thus, we draw the first claim.
Then, for the second claim, token generation in MLLMs typically uses stochastic decoding (e.g., Top $K$ sampling \cite{holtzman2019curious}) and candidate selection (e.g., beam search \cite{vinyals2015show}). Because the model attends to both foreground and background regions, these decoding steps can occasionally prefer candidates where background cues dominate. The result may be fluent yet factually incorrect outputs, i.e., hallucinations. As shown in \textbf{Figure ~\ref{fig:motivation_2}}, when asked ``Is there a traffic signal light directing traffic?'', the model shows similar gradient saliency on both the actual traffic lights and nearby façades and then confidently outputs the incorrect answer ``Yes, there are two traffic lights directing traffic''. This illustrates how spurious correlations can steer token selection and induce hallucinations.

\subsection{Motivation Analysis}
\label{sec:cc}
In this section, we revisit the mechanism by which MLLMs generate answers from a causal perspective. Then, we explore a causal-based solution helping to disentangle spurious correlations, thereby mitigating hallucinations.

We begin by constructing an SCM to formalize the data-generating process of multimodal inputs (\textbf{Figure \ref{fig:scm_mml_data}}). Let $I=(I_v,I_t)$ be the paired image-text input, $Y$ denotes the ground-truth answer. 
Following \cite{suter2019robustly,deshpande2022deep}, we assume that the input $I$ can be regarded as generated from a set of generative factors, including: (i) causal factors $L_c$ that are related to $Y$, and (ii) non-causal factors $L_s$ that reflect irrelevant influences such as background. Thus, we get $L_c \rightarrow I$ and $L_s \rightarrow I$. According to \cite{hu2022causal,deshpande2022deep}, $L_c$ corresponds to semantic attributes, e.g., the presence and category of objects. In contrast, $L_s$ corresponds to irrelevant environmental elements, e.g., background and lighting, which are not causally related to $Y$. Since both $L_c$ and $L_s$ represent high-level semantic information underlying input $I$, but only $L_c$ is causally determined by $Y$, we naturally define $Y$ as the cause of $L_c$, i.e., $Y \rightarrow L_c$. To this end, we obtain \textbf{Figure \ref{fig:scm_mml_data}}.
Next, we model the generation process of MLLMs (\textbf{Figure \ref{fig:scm_mllm_inference}}). Given an input $I$, the model generates a token sequence $o=\{o_1, \dots, o_T\}$, containing the predicted answer $\tilde{Y}=\{\tilde{y}_1,\cdots,\tilde{y}_{T_{\tilde{Y}}}\}$. 
Ideally, the model should represent only causal factors $L_c$ within $o$, excluding influence from non-causal factors $L_s$. 
In this case, $\tilde{Y}$ would depend solely on $L_c$ and remain invariant to variations in $L_s$.

In \textbf{Section \ref{sec:empirical}}, we have given our two arguments that (i) outcome-based rewards of GRPO may introduce spurious correlations, and (ii) these spurious correlations may lead to hallucinations in MLLMs. To mitigate spurious correlations, an effective method is to promote the model to use causal factors $L_c$ and exclude non-causal factors $L_s$ for token generation.
According to \cite{guo2025deepseek,shao2024deepseekmath}, the intermediate reasoning steps play a crucial role in producing the predicted answer.
Inspired by \cite{pearl2009causality,wang2025causal,yang2023invariant}, we propose to incorporate constraints that encourage preferring causally essential reasoning steps, i.e., tokens that are both sufficient and necessary. 
Specifically, sufficiency means that including these tokens improve answer accuracy; while necessity means that, in the absence of these tokens, the generated answers become inaccurate.
Thus, requiring generated tokens to satisfy causal sufficiency and necessity forces the model to produce tokens that carry indispensable and predictive information about $Y$, i.e., tokens that reflect $L_c$.
Consequently, the background-driven tokens will fail to meet these criteria and thus cannot sustain the spurious path $L_s \to \tilde{Y}$, thus reducing hallucinations.
Here, we provide the concept of the probability of causal sufficiency and necessity (PNS) in MLLMs:
\begin{definition}[PNS for MLLMs]\label{def:cc}
Let $I=(I_v,I_t)$ be an input image-text pair and $o = \{o_1,\dots,o_T\}$ denotes the output token sequence generated by the MLLM. Let $\tilde{Y}$ denote the predicted answer within $o$ and $Y$ the ground-truth answer. The probability that $o$ satisfies causal completeness for the model prediction can be defined as:
\begin{equation}\label{eq:def_cc}
\begin{aligned}
    \scalebox{0.93}{$C(o)=P_{\rm sufficiency}P(\tilde{Y}\neq Y, \bar{o}, I)
    +P_{\rm necessity}P(\tilde{Y}= Y, o, I),$}
\end{aligned}
\end{equation}
where $P_{\rm sufficiency}=P(\tilde{Y}_{{\rm do}(o)}=Y| \tilde{Y}\neq Y, \bar{o},I)$ is for causal sufficiency: the probability that the model answer becomes correct ($\tilde{Y}=Y$) when we intervene ${\rm do}(o)$, given that under $\bar{o}$ the answer is incorrect ($\tilde{Y}\neq Y$). 
Similarly, $P_{\rm necessity}=P(\tilde{Y}_{{\rm do}(o_{<t},\bar{o}_t,o_{>t})}\neq Y| \tilde{Y}=Y,o,I)$ is for causal necessity: the probability that the answer becomes incorrect ($\tilde{Y}\neq Y$) when we intervene ${\rm do}(o_{<t},\bar{o}_t,o_{>t})$, given that $o$ yields the correct answer ($\tilde{Y}=Y$).
\end{definition}
\textbf{Definition~\ref{def:cc}} implies that a sequence $o$ with a larger $C(o)$ contains more causally relevant tokens and is therefore more likely to yield a correct answer.
Concretely, the sufficiency term $P_{\rm sufficiency}$ measures the probability that enforcing $o$ (via $\mathrm{do}(o)$) turns an otherwise incorrect output into a correct one, i.e., $o$ provides sufficient causal evidence. The necessity term $P_{\rm necessity}$ measures the probability that perturbing a token in an otherwise-correct sequence renders the answer incorrect, i.e., that token is necessary for the correct outcome. 
Together, these complementary tests strengthen the true causal signal $L_c$ while suppressing spurious, non-causal distractions $L_s$ that drive hallucinations. 
\section{Methods}
\label{sec:methods}

\begin{figure*}[t]
    \centering
    \includegraphics[width=\textwidth]{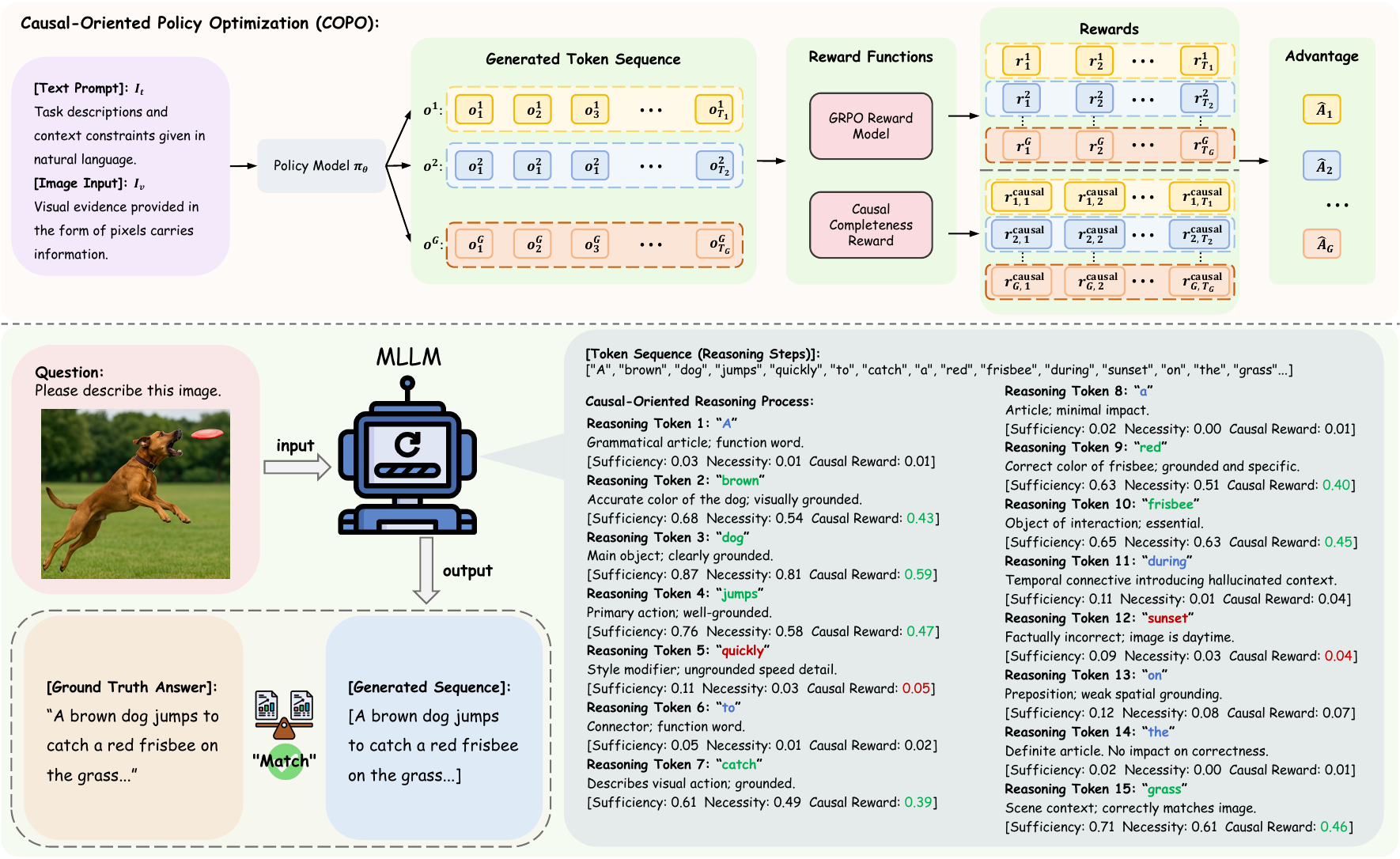}
    \caption{Overview of our causal-oriented policy optimization framework. 
    The upper part is the pipeline of COPO, and the lower part is the calculation process of our proposed causal completeness rewards.
    }
    \label{fig:overview}
\end{figure*}

Based on the above discussion, we propose COPO, a causal-oriented policy optimization framework to mitigate hallucinations. 
It constrains the model to prefer tokens (i.e., reasoning steps) with stronger causal completeness to suppress spurious correlations, thereby mitigating hallucinations.
Specifically, we first define a causal completeness reward that quantifies each token's contribution to the correctness by jointly considering its causal sufficiency and necessity. This reward is then integrated into a modified advantage function. The resulting advantage is incorporated into GRPO, guiding the model to attend to causal complete tokens for accurate answers. The framework of COPO is shown in \textbf{Figure \ref{fig:overview}}, with the pseudo-code in \textbf{Appendix \ref{sec_app:pseudo_code}}.

\subsection{Causal Completemess Reward}
\label{sec:causal_completeness_reward}
In this subsection, we introduce the proposed causal completeness reward function.
It integrates two complementary constraints: (i) a causal sufficiency constraint encourages the model to focus on tokens that individually contribute to drive correct answers; and (ii) a causal necessity constraint measures the degree to which each token is indispensable for maintaining the correct answer. 
We first provide the formalization of the causal sufficiency score and necessity scores. Then, we combine these two using a weighted convex combination to produce a causal completeness reward that considers both causal sufficiency and necessity. 

\subsubsection{Causal Sufficiency Score}
\label{sec:css}
Denote the output token sequence by $o=\{o_1,\cdots,o_T\}$, consisting of predicted answer $\tilde{Y}$ and reasoning tokens $\bar{o}=\{o\}\setminus \{\tilde{Y}\}=\{\bar{o}_1,\cdots,\bar{o}_{T_{\bar{o}}}\}$. 
Based on \textbf{Section \ref{sec:cc}}, reasoning tokens $\bar{o}$ have a significant impact on the correctness of the answer $\tilde{Y}$. Therefore, we hope to impose constraints on the reasoning steps to induce the model to generate the correct answer.
For a given token $\bar{o}_t$, we construct an intervention by masking subsequent tokens $\bar{o}_{>t}$ while retaining the sequence $\bar{o}_{\leq t}$. 
We input $\bar{o}_{\leq t}$ and $\bar{o}_{< t}$ into the model, and obtain $\{o^{k}_{(t)}\}_{k=1}^H=\{\bar{o}^k_1,\cdots,\bar{o}^k_t,\tilde{o}^k_{t+1},\cdots,\tilde{o}^k_{T_k}\}_{k=1}^H$ and $\{o^{k}_{(t-1)}\}_{k=1}^H=\{\bar{o}^k_1,\cdots,\bar{o}^k_{t-1},\tilde{o}^k_{t},\cdots,\tilde{o}^k_{T_k}\}_{k=1}^H$, where $\tilde{o}_{(\cdot)}^{(\cdot)}$ denotes the newly generated token. Correspondingly, we receive $H$ answers $\{\tilde{Y}^k_{(t)}\}_{k=1}^H$ and $\{\tilde{Y}^k_{(t-1)}\}_{k=1}^H$. 
To evaluate the contribution of $\bar{o}_t$, we compare the reward with and without $\bar{o}_t$, averaging across the $H$ sampled continuations to account for decoding stochasticity \cite{shi2024thorough,herrera2025overview} and thereby estimate the token's expected importance.
The causal sufficiency score can be expressed as:
\begin{equation}\label{eq:suff_score}
\begin{aligned}
    S_{\text{suff}}(\bar{o}_t) =& \big(\frac{1}{H} \sum_{k=1}^H r(\tilde{Y}^k_{(t)})-\frac{1}{H} \sum_{k=1}^H r(\tilde{Y}^k_{(t-1)})\big) \\&\cdot\mathbb{I}(r(\tilde{Y}^k_{(t)})>r(\tilde{Y}^k_{(t-1)})),
\end{aligned}
\end{equation}
where $\mathbb{I}(r(\tilde{Y}^k_{(t)})>r(\tilde{Y}^k_{(t-1)}))$ is an indicator function that takes 1 when $r(\tilde{Y}^k_{(t)})>r(\tilde{Y}^k_{(t-1)})$, and 0 otherwise.

\textbf{Eq.\ref{eq:suff_score}} measures whether incorporating the current token improves the model's reasoning by yielding a higher reward compared to the reasoning before its inclusion.
It reflects the degree to which the inclusion of the token improves the correctness of the answer, which aligns with the causal sufficiency of \textbf{Definition \ref{def:cc}}. 
A higher $S_{\text{suff}}(\bar{o}_t)$ indicates that $\bar{o}_t$ is essential to trigger the correct answer.

\subsubsection{Causal Necessity Score}
\label{sec:cns}
To calculate the causal necessity score, we approximate the counterfactual intervention by masking.
Similarly, let a token sequence be $o = \{o_1, \dots, o_T\}$, composed of answer $\tilde{Y}$ and reasoning tokens $\bar{o}=\{o\}\setminus \{\tilde{Y}\}$. Given $\bar{o}_t$, we define the counterfactual reasoning token sequence $\bar{o}^{\rm mask}_{(t)} = \{\bar{o}_1, \dots,\bar{o}^{\rm mask}_t,\cdots, \bar{o}_{T_{\bar{o}}}\}$ where token $\bar{o}_t$ is replaced with a masked token $\bar{o}^{\rm mask}_t$, i.e., set the value of $\bar{o}_t$ to zero. 
Then, we receive $\tilde{Y}^{\rm mask}_{(t)}$. The causal necessity score can be expressed as:
\begin{equation}\label{eq:nec_score}
    S_{\text{nec}}(\bar{o}_t) = r(\tilde{Y})-r(\tilde{Y}^{\rm mask}_{(t)}).
\end{equation}
\textbf{Eq. \ref{eq:nec_score}} estimates whether excluding the current token results in a worse answer compared to when the token is included (i.e., whether the reward of the former decreases). This reflects the extent to which the answer becomes less accurate when the token is excluded, which aligns with the causal necessity of \textbf{Definition \ref{def:cc}}. A higher $S_{\text{nec}}(\bar{o}_t)$ indicates that the token may have a substantial influence on the correctness of the answer, i.e., removing the influence of this token will disrupt the answer. In particular, we do not follow the multiple sampling and averaging process in \textbf{Eq.\ref{eq:suff_score}}, since most existing MLLMs have considered decoding stochasticty when generating the answer $\tilde{Y}$ from reference process $o$, such as Top $K$ \cite{holtzman2019curious} and Beam Search \cite{vinyals2015show}.

\subsubsection{Causal Completeness Reward}
\label{sec:ccr}
To identify tokens that are both causally sufficient and necessary, it is essential to jointly evaluate their contributions. Accordingly, we combine the sufficiency and necessity scores into a causal completeness reward.

Specifically, for each token $\bar{o}_t$, we have obtained its causal sufficiency score $S_{\text{suff}}(\bar{o}_t)$ and necessity score $S_{\text{nec}}(\bar{o}_t)$, both normalized to the range $[0,1]$. We combine the two scores into a unified causal completeness reward:
\begin{equation}\label{eq:causal_reward}
    r_{\text{causal}}(\bar{o}_t) = \lambda_{\rm s} \cdot S_{\text{suff}}(\bar{o}_t) + \lambda_{\rm n} \cdot S_{\text{nec}}(\bar{o}_t),
\end{equation}
where $\lambda_{\rm s},\lambda_{\rm n}\in[0,1]$ are weighting coefficients. 
This reward is subsequently incorporated into the GRPO objective, enabling the model to explicitly prioritize causally complete tokens, thereby suppressing hallucinations.

\begin{table*}[t]
    \centering          
    \caption{Results on CHAIR and POPE (F1 score) hallucination evaluation. The best results are highlighted in \textbf{bold}. }
    \resizebox*{\linewidth}{!}{
    \begin{tabular}{@{}lcccccccccccc@{}}           
    \toprule          
    \multirow{2}{*}{\textbf{Method}} &\multicolumn{3}{c}{\textbf{InstructBLIP}} &\multicolumn{3}{c}{\textbf{MiniGPT-4}} & \multicolumn{3}{c}{\textbf{LLaVA-1.5}} & \multicolumn{3}{c}{\textbf{Qwen-VL}}\\    
    \cmidrule(lr){2-4} \cmidrule(lr){5-7} \cmidrule(lr){8-10} \cmidrule(lr){11-13}
    & $\mathrm{CHAIR_S}\downarrow$ & $\mathrm{CHAIR_I}\downarrow$ & POPE $\uparrow$ & $\mathrm{CHAIR_S}\downarrow$ & $\mathrm{CHAIR_I}\downarrow$ & POPE $\uparrow$ & $\mathrm{CHAIR_S}\downarrow$ & $\mathrm{CHAIR_I}\downarrow$ & POPE $\uparrow$ & $\mathrm{CHAIR_S}\downarrow$ & $\mathrm{CHAIR_I}\downarrow$ & POPE $\uparrow$ \\            
    \midrule            
    Vanilla & 58.8 & 23.7 & 80.0 & 31.8 & 9.9 & 58.5 & 45.0 & 14.7 & 82.2 & 46.0 & 12.5 & 85.2\\      
    Beam Search \cite{vinyals2015show} & 55.6 & 15.8 & 84.4 & 30.6 & 9.5 & 70.3 & 48.8 & 13.9 & 84.9 & 41.8 & 10.8 & 85.3 \\
    Nucleus \cite{holtzman2019curious} & 54.6 & 24.8 & 79.8 & 32.6 & 10.7 & 52.8 & 48.8 & 14.2 & 83.1 & 49.2 & 13.1 & 84.5\\
    DoLa \cite{chuang2023dola} & 48.4 & 15.9 & 83.4 & 32.2 & 10.0 & 72.8 & 47.8 & 13.8 & 83.2 & 46.8 & 12.9 & 85.8\\            
    OPERA \cite{huang2024opera} & 46.4 & 14.2 & 84.8 & 26.2 & 9.5 & 73.3 & 44.6 & 12.8 & 85.4 & 44.6 & 11.4 & 86.1\\             
    VCD \cite{leng2024mitigating} & 58.0 & 17.0 & 79.9 & 33.8 & 11.1 & 56.0 & 54.0 & 16.0 & 83.1 & 46.4 & 11.9 & 84.7\\
    DeCo \cite{wang2025mllmseedynamiccorrection} & 41.2 & 14.4 & 84.9 & 27.0 & 8.8 & 77.4 & 37.8 & 11.1 & 86.7 & 42.2 & 10.7 & 86.3 \\
    VTI \cite{liu2025reducing} & 43.4 & 11.8 & 83.2 & - & - & - & 35.8 & 11.1 & 85.9 & - & - & 85.2 \\
    HA-DPO \cite{zhao2023beyond} & 43.5 & 15.2 & 82.5 & 27.8 & 9.5 & 76.4 & 38.2 & 11.0 & 82.7 & 43.2 & 11.1 & - \\
    CLIP (OHD) \cite{liu2024investigating} & 47.2 & 15.0 & 84.2 & 28.2 & 9.3 & 75.7 & 45.2 & 14.1 & 84.9 & 45.5 & 11.9 & 86.1 \\
    POVID \cite{zhou2024aligning} & 40.7 & 10.8 & 84.9 & 37.8 & 8.8 & 77.2 & 35.2 & 8.3 & 86.9 & 32.1 & 7.7 & 86.2 \\
    CSR \cite{zhou2024calibrated} & 37.6 & 9.7 & 85.5 & 22.8 & 7.7 & 77.8 & 21.0 & 6.0 & 87.0 & 20.5 & 5.8 & 86.8 \\
    GCPO \cite{gu2025group} & 38.2 & 9.3 & 85.9 & 21.9 & 8.0 & 78.6 & 21.5 & 5.8 & 87.2 & 19.9 & 6.0 & 86.5 \\
    \midrule
    \textbf{Ours} (\textit{vs} SOTA) & \textbf{35.9} ($\downarrow$ 1.7) & \textbf{8.6} ($\downarrow$ 0.7) & \textbf{86.9} ($\uparrow$ 1.0) & \textbf{20.6} ($\downarrow$ 1.3) & \textbf{7.1} ($\downarrow$ 0.6) & \textbf{80.1} ($\uparrow$ 1.5) & \textbf{19.8} ($\downarrow$ 1.2) & \textbf{5.3} ($\downarrow$ 0.5) & \textbf{88.0} ($\uparrow$ 0.8) & \textbf{18.8} ($\downarrow$ 1.1) & \textbf{5.2} ($\downarrow$ 0.6) & \textbf{88.3} ($\uparrow$ 1.5)\\
    \bottomrule          
    \end{tabular}}
    \label{tab:chair}
\end{table*}

\subsection{Causal-Oriented Policy Optimization}
\label{sec:objective}
In this subsection, we propose COPO framework that extends the GRPO method \cite{shao2024deepseekmath} by incorporating the above causal completeness reward. Specifically, we first introduce a causal-oriented advantage function that integrates token-level causal completeness. We then present a joint optimization objective that modifies the GRPO policy update to prioritize causally informative tokens. It encourages the model to attend to tokens that are both causally sufficient and necessary for correct answers to reduce hallucinations.

\subsubsection{Causal-Oriented Advantage Function}
\label{sec:causal_advantage_fuction}
Following GRPO \cite{shao2024deepseekmath}, each input $I^{(n)}$ is associated with a sampled trajectories $\{o^{i} = (o^{i}_{1}, \dots, o^{i}_{T_i})\}_{i=1}^G$, including answers $\{\tilde{Y}^{i} = (\tilde{y}^{i}_{1}, \dots, \tilde{y}^{i}_{T_{Y^i}})\}_{i=1}^G$.
According to \cite{shao2024deepseekmath}, the original advantage of $o^i_t$ is calculated via \textbf{Eq.\ref{eq:adv}}, i.e., $A^{\rm orig}_{i,t}=A_i$.
Following \textbf{Eq.\ref{eq:causal_reward}}, the causal completeness reward $r^{\rm causal}_{i,t} = r_{\rm causal}(\bar{o}^{i}_{t})$ is assigned to each reasoning token by the reward function $r_{\rm causal}(\cdot)$. 
Here, $r^{\rm causal}_{i,t}$ measures the causal contribution of the $t^{th}$ reasoning token in the $i^{th}$ trajectory. For the advantage of the answer, we still follow the calculation in \textbf{Eq.\ref{eq:adv}}.
Then, to integrate causal constraints, we add the causal completeness rewards into the advantage for reasoning tokens, which can be expressed as:
\begin{equation}\label{eq:adv_causal}
\hat{A}_{i,t} = A^{\rm orig}_{i,t} + r^{\rm causal}_{i,t},\quad \text{s.t.}\; o_t^i\in\bar{o}^i,
\end{equation}
Notably, the hyperparameters $\lambda_{\rm s}$ and $\lambda_{\rm n}$ in \textbf{Eq.\ref{eq:causal_reward}} are adjusted, not only controlling the respective proportions of sufficiency and necessity, but also controlling the intensity of causal completeness rewards in the advantage (with details in \textbf{Section \ref{sec:exp_ablation_study}}). \textbf{Eq.\ref{eq:adv_causal}} encourages the model to prefer tokens with stronger causal completeness to suppress spurious correlations, mitigating hallucinations.

\begin{figure}
    \centering
    \includegraphics[width=\columnwidth]{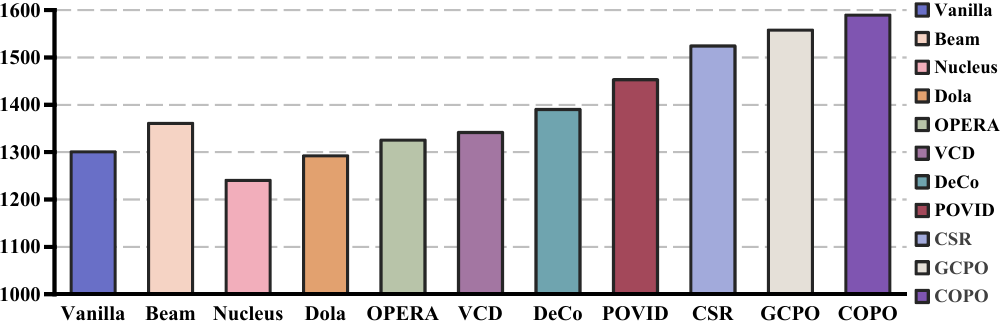}
    \caption{Results on text quality evaluation.}
    \vspace{-0.05in}
    \label{fig:mme}
    \vspace{-0.15in}
\end{figure}

\subsubsection{Overall Optimization}
\label{sec:joint_optimizaition_objective}
Based on the above causal-oriented advantage function $\hat{A}_{i,t}$, we construct the joint optimization process. Let $\pi_\theta(o^{i}_{t} | I^{(n)},o^i_{<t})$ denote the probability of generating the token $o^{i}_{t}$, and let the corresponding probability under the old policy be $\pi_{\theta_{\text{old}}}(o^{i}_{t} | I^{(n)},o^i_{<t})$. The importance weight is defined as:
\begin{equation}
    \rho_{i,t} = \frac{\pi_\theta(o^{i}_{t} \mid I^{(n)},o^i_{<t})}{\pi_{\theta_{\text{old}}}(o^{i}_{t} \mid I^{(n)},o^i_{<t})}.
\end{equation}
The joint objective function becomes:
\begin{equation}\label{eq:grpo_objective}
\begin{aligned}
    &\mathcal{J}_{\rm COPO}(\theta) = \mathbb{E}_{q \sim P,\, \{o^{i}\} \sim \pi_{\theta_{\text{old}}}}\\
    &\left[\frac{1}{G} \sum\limits_{i=1}^G \frac{1}{T_i}\sum\limits_{t=1}^{T_i}\bigg(\min (\rho_{i,t} \cdot \hat{A}_{i,t},\Psi(\hat{A}_{i,t})) - \beta\, \mu(\pi_\theta)\bigg)\right]\\
    &\text{s.t.}\, \Psi(\hat{A}_{i,t})=\mathrm{clip}(\rho_{i,t}, 1 - \epsilon, 1 + \epsilon) \cdot \hat{A}_{i,t},\\
    &\qquad   \mu(\pi_\theta)=D_{\mathrm{KL}}\left( \pi_\theta \Vert \pi_{\text{ref}} \right),
\end{aligned}
\end{equation}
where $\epsilon$ is the clipping threshold controlling update range, $\beta$ is the coefficient for the KL penalty, $\pi_{\text{ref}}$ is a reference policy, $P$ is the distribution of query, and $D_{\mathrm{KL}}(\cdot\|\cdot)$ denotes KL divergence.
This objective preserves the stability and contrastiveness of GRPO while explicitly enhancing policy learning with token-level causal supervision. As a result, we encourage the model to prioritize causally sufficient and necessary tokens, reducing hallucinations in MLLMs.

\begin{table}[t]
    \centering
    \caption{Results on GPT-4 assisted evaluation.}
    \resizebox*{\columnwidth}{!}{
    \begin{tabular}{l | c c c } 
    \toprule
    \multirow{2}{*}{\textbf{Method}} & \multicolumn{3}{c}{\textbf{GPT-4 Assisted Evaluation}} \\
    & Accuracy (A) & Correctness (C) & Detailedness (D) \\
    \midrule
    Vanilla & 5.21 & 6.31 & 8.18 \\
    Beam Search \cite{vinyals2015show} & - & 5.53 & 5.15 \\
    OPERA~\cite{huang2024opera}  & - & 6.32 & 5.16 \\
    VCD~\cite{leng2024mitigating}  & 4.15 & - & 3.85 \\
    DeCo \cite{wang2025mllmseedynamiccorrection} & 7.42& 6.25 & 7.96 \\
    \midrule
    \textbf{Ours} (\textit{vs} SOTA) & \textbf{8.71} ($\uparrow1.29$) & \textbf{6.89} ($\uparrow0.57$) & \textbf{9.58} ($\uparrow1.40$) \\
    \bottomrule
    \end{tabular}}
    \label{tab:tq_gpt4}
\end{table}

\begin{table}[t]
    \centering
    \caption{The effect of causal completeness components.}
    \resizebox*{\columnwidth}{!}{
    \begin{tabular}{l | c c c | c  | c c c  } 
    \toprule
    \multirow{2}{*}{\textbf{Methods}} & \multicolumn{3}{c|}{\textbf{Hallucination Resistance}} & {\textbf{Text Quality}} & \multicolumn{3}{c}{\textbf{GPT-4 Assistance}}\\
    & $\mathrm{CHAIR_S}$ & $\mathrm{CHAIR_I}$ & POPE & MME & A & C & D \\
    \midrule
    COPO w/o $S_{\rm suff}$ & 21.7 & 7.5 & 86.7 & 1531.5 & 7.78 & 6.21 & 8.89 \\
    COPO w/o $S_{\rm nec}$  & 22.5 & 6.9 & 87.0 & 1522.3 & 7.81 & 6.18 & 8.75 \\
    COPO w/o $S_{\rm suff}$\&$S_{\rm nec}$  & 30.5 & 10.9 & 85.9 & 1489.2 & 7.21 & 5.77 & 8.19 \\
    \midrule
    COPO  & 19.8 & 5.3 & 88.0 & 1589.3 & 8.71 & 6.89 & 9.58 \\
    \bottomrule
    \end{tabular}}
    \label{tab:exp_abla}
    \vspace{-0.1in}
\end{table}
\section{Experiments}
\label{sec:experiments}
In this section, we conduct extensive experiments on multiple reasoning benchmarks to verify the effectiveness of our method. More details, experiments, and analysis are provided in the \textbf{Appendix} due to space limitations. 

\subsection{Experimental Setup}
\label{sec:exp_setup}

\textbf{Baselines and Benchmarks.} 
We select a series of strong baselines for evaluation, including DoLa \cite{chuang2023dola}, OPERA \cite{huang2024opera}, VCD \cite{leng2024mitigating}, DeCo \cite{wang2025mllmseedynamiccorrection}, VTI \cite{liu2025reducing}, HA-DPO \cite{liu2023mitigating}, CLIP (OHD) \cite{liu2024investigating}, POVID \cite{zhou2024aligning}, CSR \cite{zhou2024calibrated}, and GCPO \cite{gu2025group}.
We evaluate our method using a diverse suite of benchmarks, including CHAIR \cite{rohrbach2018object}, POPE \cite{li2023evaluating},  and MME \cite{fu2024mmecomprehensiveevaluationbenchmark}. We also conduct experiments on $V^*$ Bench\cite{wu2024v}, HR-Bench(4K/8K) \cite{wang2025divide}, refCOCO \cite{caesar2018coco}, refCOCO+ \cite{caesar2018coco}, refCOCOg \cite{kazemzadeh2014referitgame}, ReasonSeg \cite{lai2024lisa}, MathVista \cite{lu2023mathvista}, MathVerse \cite{zhang2024mathverse}, MathVision \cite{wang2024measuring}, WeMath \cite{qiao2024we}, DynaMath \cite{zou2024dynamath}, and LogicVista \cite{xiao2024logicvista} in the \textbf{Appendix}. For GPT-4-assisted evaluation, we compare model-generated captions on images, with GPT-4o rating on accuracy (A), correctness (C), and detailedness (D).

\noindent\textbf{Implementation Details.} 
We adopt four representative MLLMs for evaluation: InstructBLIP \cite{dai2023instructblipgeneralpurposevisionlanguagemodels}, MiniGPT-4 \cite{zhu2023minigpt}, LLaVA-1.5 \cite{liu2024improved}, and Qwen-VL \cite{bai2023qwenvlversatilevisionlanguagemodel}. We build our method on top of a pre-trained 7B-sized model, and fine-tune it using our proposed COPO framework on A100 GPU clusters. More details are shown in the \textbf{Appendix}.

\subsection{Main Results}
\label{sec:exp_main_results}

\noindent\textbf{Hallucination Assessment.} 
CHAIR measures object hallucinations in captioning by computing instance-level ($\mathrm{CHAIR_I}$) and sentence-level ($\mathrm{CHAIR_S}$) error rates, while POPE evaluates object-level hallucinations via a probing-based VQA setting. As shown in \textbf{Table \ref{tab:chair}}, our COPO yields lower hallucination rates in CHAIR and higher F1 scores in POPE, demonstrating the effectiveness of our approach.

\noindent\textbf{Performance Evaluation.}
MME \cite{fu2024mmecomprehensiveevaluationbenchmark} further evaluates the capability in producing high-quality language outputs. These benchmarks assess accuracy, relevance, and the consistency of language grounding. As shown in \textbf{Figure \ref{fig:mme}}, our method delivers strong performance on both benchmarks, surpassing baselines and demonstrating the advantages.

\noindent\textbf{GPT-4 Assisted Evaluation.}
To assess the perceived quality of generated captions, we adopt GPT-4o as an automatic evaluator following \cite{huang2024opera, leng2024mitigating}. As shown in \textbf{Table~\ref{tab:tq_gpt4}}, our method achieves substantial gains, notably reaching the highest scores in both accuracy and content richness.

\begin{figure}
    \centering
    \includegraphics[width=\columnwidth]{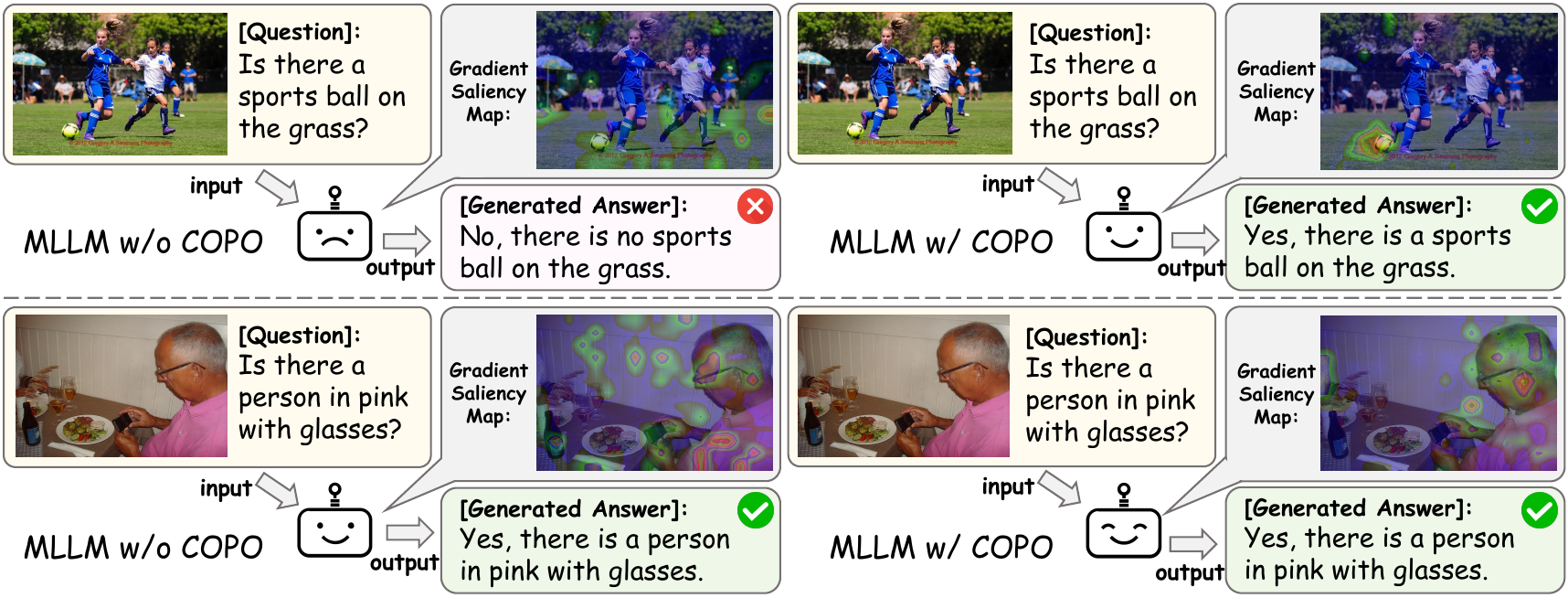}
    \caption{The effect of causal completeness rewards.}
    \label{fig:exp_compare}
\end{figure}

\begin{figure}[t]
    \centering
    \begin{subfigure}[t]{0.23\textwidth}
        \centering
        \includegraphics[width=0.9\textwidth]{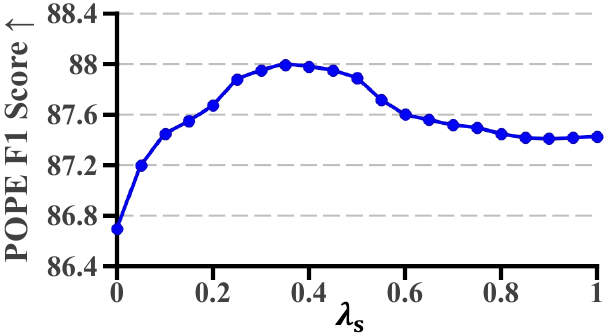}
        \caption{$\lambda_s$ v.s. POPE F1 Score}
        \label{fig:alba1}
    \end{subfigure}
    \hfill
    \begin{subfigure}[t]{0.23\textwidth}
        \centering
        \includegraphics[width=0.9\textwidth]{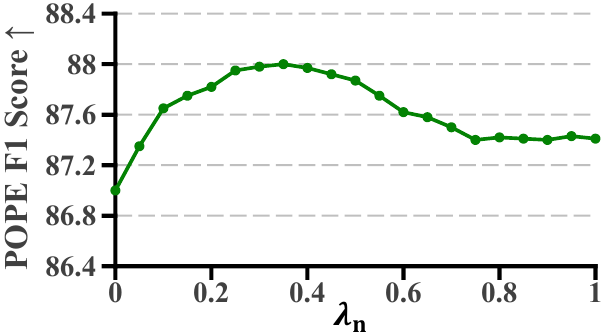}
        \caption{$\lambda_n$ v.s. POPE F1 Score}
        \label{fig:alba2}
    \end{subfigure}
    \caption{Parameter sensitivity.}
    \label{fig:exp_alba}
\end{figure}

\subsection{Ablation Study}
\label{sec:exp_ablation_study}

\noindent\textbf{The effect of different components.}
We examine three ablated variants: (i) COPO w/o $S_{\rm suff}$; (ii) COPO w/o $S_{\rm nec}$; (iii) COPO w/o $S_{\rm suff}$\&$S_{\rm nec}$.
The results in \textbf{Table \ref{tab:exp_abla}} demonstrate the effectiveness of each component within COPO.

\noindent\textbf{Parameter sensitivity.}
We conduct experiments on the hyperparameters $\lambda_{{s}}$ and $\lambda_{{n}}$. We search $\lambda_{{s}}$ and $\lambda_{{n}}$ over $[0, 1]$. 
\textbf{Figure~\ref{fig:exp_alba}} shows that the optimal result is at $\lambda_{{s}} = 0.35$ and $\lambda_{{n}} = 0.35$, which are our final configuration.
\section{Conclusion}
\label{sec:conclusion}
In this paper, we explore the hallucinations in MLLMs and propose an effective method to address them. Through empirical and causal analyses, we reveal that (i) MLLMs may allocate disproportionately more gradient saliency to task-irrelevant background regions than text-only LLMs, indicating spurious correlations; (ii) GRPO's outcome-only rewards can inadvertently reinforce such spurious correlations, increasing the risk of hallucinations during generation and candidate selection. 
To address this, we propose Causal-Oriented Policy Optimization (COPO), which injects token-level causal sufficiency and necessity constraints via a causal-completeness reward into the GRPO advantage. COPO steers learning toward evidence-grounded tokens and away from background-dependent shortcuts. Experiments conducted on various benchmarks demonstrate the advantages of COPO in reducing hallucinations.

{
    \small
    \bibliographystyle{ieeenat_fullname}
    \bibliography{main}
}

\clearpage
\appendix
\setcounter{page}{1}
\maketitlesupplementary

\section*{Appendix}
\label{sec:appendix}
The appendix is organized into several sections:

\begin{itemize}
    \item \textbf{Appendix \ref{sec_app:notation}} provides details for all notations used in this paper.
    \item \textbf{Appendix \ref{sec_app:pseudo_code}} provides the pseudo-code of our method.
    \item \textbf{Appendix \ref{sec_app:discussion}} provides more discussion about the proposed methodology.
    \item \textbf{Appendix \ref{sec_app:dataset}} provides details for the datasets used in this paper.
    \item \textbf{Appendix \ref{sec_app:baseline}} provides details for the baselines mentioned in the main text.
    \item \textbf{Appendix \ref{sec_app:benchmark}} provides details for the benchmarks used in the experiments.
    \item \textbf{Appendix \ref{sec_app:implementation}} contains details for the implementation of the experiment.
    \item \textbf{Appendix \ref{sec_app:experiment}} provides the full results and analyses of the experiment.
\end{itemize}

\section{Notations}
\label{sec_app:notation}
In this section, we briefly describe the symbols that we mainly use in this article. In \textbf{table \ref{tab:notation}}, we give the definitions of notation according to their role.

\begin{table*}
    \centering
    \renewcommand{\arraystretch}{1.1}
    \setlength{\tabcolsep}{1mm}
    \begin{tabular}{cc}
    \toprule
    Notations & Definitions\\
    \midrule
    \textit{Notations of Data} & \textit{Definitions of Data} \\
    \midrule
    $I^{(n)}=\{I_v^{(n)},I_t^{(n)}\}$ & The paird input with a image $I_v^{(n)}$ and a text prompt $I_t^{(n)}$\\
    $\mathcal{D}=\{(I_v^{(n)},I_t^{(n)},Y^{(n)})\}_{n=1}^N$ & The dataset of paired inputs and corresponding labels with $N$ samples \\
    \midrule
    \textit{Notations of Model} & \textit{Definitions of Model} \\
    \midrule
    $S_{\text{suff}}(\cdot)$ & Causal sufficiency score \\
    $S_{\text{nec}}(\cdot)$ & Causal necessity score \\
    $r_{\text{causal}}(\cdot)$ & Causal completeness reward function \\
    $r(\cdot)$ & Reward model in GRPO \\
    $\pi_\theta(o^{i}_{t} | I^{(n)},o^i_{<t})$ & Probability of generating token $o^{i}_{t}$  \\
    $\pi_{\theta_{\text{old}}}(o^{i}_{t} | I^{(n)},o^i_{<t})$ & Probability generating token $o^{i}_{t}$ under the old policy \\
    $\pi_{\theta_{\text{ref}}}$ & Reference policy \\
    \midrule
    \textit{Notations of Variables} & \textit{Definitions of Variables} \\
    \midrule
    $I_v$ & Image input \\
    $I_t$ & Text prompt \\
    $Y$ & Corresponding ground-truth answer \\
    $L_c$ & Set of causal factors \\
    $L_s$ & Set of non-causal factors \\
    $o=\{o_t\}_{t=1}^T$ & Token sequence generated from MLLM with $T$ tokens\\
    $\tilde{Y}=\{\tilde{y}_1,\cdots,\tilde{y}_{T_{y^i}}\}$ & The final answer within $o$ \\
    $\bar{o}=\{o\}\setminus \{\tilde{Y}\}=\{\bar{o}_1,\cdots,\bar{o}_{T_{\bar{o}}}\}$ & Reasoning tokens of reasoning steps \\
    $\{o^{k}_{(t)}\}_{k=1}^H=\{\bar{o}^k_1,\cdots,\bar{o}^k_t,\tilde{o}^k_{t+1},\cdots,\tilde{o}^k_{T_k}\}_{k=1}^H$ & Token sequences generated by MLLM with $\bar{o}_{\leq t}$ \\
    $\tilde{o}_{(\cdot)}^{(\cdot)}$ & Tokens which are newly generated \\
    $\bar{o}^{\rm mask}_{(t)} = \{\bar{o}_1, \dots,\bar{o}^{\rm mask}_t,\cdots, \bar{o}_{T_{\bar{o}}}\}$ & Token sequence with token $o_t$ is masked \\
    $\lambda_s,\lambda_n$ & Hyperparameters for causal sufficiency score and causal necessity score \\
    $\{ o^{i} = (o^{i}_{1}, \dots, o^{i}_{T_i}) \}_{i=1}^G$ & Group of $G$ sampled trajectories \\
    $r_t^i$ & Reward of the $t^{th}$ token in the $i^{th}$ trajectory \\
    $r^{\text{causal}}_{i,t}$ & Causal completeness reward of the $t^{th}$ token in the $i^{th}$ trajectory \\
    $A^{\text{orig}}_{i,t},\; A_{i,t}$ & The GRPO advantage function \\
    $\hat{A}_{i,t}$ & The causal-oriented advantage function \\
    $\rho_{i,t},\; R_{i,j}(\theta)$ & Importance weight of the $t^{th}$ token in the $i^{th}$ trajectory \\
    \midrule
    \textit{Notations of Learning Objective} & \textit{Definitions of Learning Objective} \\
    \midrule
    $\mathcal{J}_{\rm GRPO}(\theta)$ & The optimization objective of GRPO \\
    $\mathcal{J}_{\rm COPO}(\theta)$ & The optimization objective of COPO \\
    $\Psi(\hat{A}_{i,t})=\mathrm{clip}(\rho_{i,t}, 1 - \epsilon, 1 + \epsilon) \cdot \hat{A}_{i,t}$ & The clipped surrogate objective \\
    $\mu(\pi_\theta)=D_{\mathrm{KL}}\left( \pi_\theta \parallel \pi_{\text{ref}} \right)$ & The KL divergence between $\pi_\theta$ and $\pi_{\text{ref}}$ \\
    \bottomrule
    \end{tabular}
    \caption{The definitions of notations}
    \label{tab:notation}
\end{table*}

\section{Pseudo-code}
\label{sec_app:pseudo_code}
The pseudo-code of causal-oriented reinforcement learning framework is shown in \textbf{Algorithm \ref{alg:causal-reward}} and \textbf{Algorithm \ref{alg:grpo-objective}}, mainly showing the steps on the basis of the implementation of base models. The main code is provided in the supplementary materials.

\begin{algorithm}[h]
\caption{Causal Completeness Reward}
\label{alg:causal-reward}
\textbf{Input}: Token sequence $\mathbf{o} = \{o_1, o_2, \ldots, o_T\}$, ground-truth answer $Y$, weights $\lambda_s$, $\lambda_n$ \\
\textbf{Output}: Token-level causal reward $\mathbf{r}_\text{causal} = \{r_\text{causal}(o_t)\}_{t=1}^T$
\begin{algorithmic}[1]
\FOR{$t = 1$ \textbf{to} $T$}
    \STATE \textit{// Causal sufficiency}
    \STATE Generate answers $\{\tilde{Y}^k_{(t)}\}_{k=1}^H$ from prefix $\mathbf{o}_{<t}$
    \STATE Generate answers $\{\tilde{Y}^k_{(t-1)}\}_{k=1}^H$ from prefix $\mathbf{o}_{< t}$
    \IF{$r(\tilde{Y}^k_{(t)})>r(\tilde{Y}^k_{(t-1)})$}
        \STATE $S_\text{suff}(o_t) \leftarrow  \frac{1}{H} \sum_{k=1}^H r(\tilde{Y}^k_{(t)})-\frac{1}{H} \sum_{k=1}^H r(\tilde{Y}^k_{(t-1)})$
    \ELSE
        \STATE $S_\text{suff}(o_t) \leftarrow 0$
    \ENDIF

    \STATE \textit{// Causal necessity}
    \STATE Construct a masked sequence where $o_t$ is replaced by a token $\tilde{o}_t$
    \STATE Generate answer $\tilde{Y}^{\rm mask}_{(t)}$ from the masked sequence
    \STATE $S_\text{nec}(o_t) \leftarrow r(\tilde{Y})-r(\tilde{Y}^{\rm mask}_{(t)})$

    \STATE \textit{// Causal completeness reward}
    \STATE $r_\text{causal}(o_t) \leftarrow \lambda_s S_\text{suff}(o_t) + \lambda_n S_\text{nec}(o_t)$
\ENDFOR
\STATE \textbf{return} $\{r_\text{causal}(o_t)\}_{t=1}^T$
\end{algorithmic}
\end{algorithm}

\begin{algorithm}[h]
\caption{Causal-Oriented Reinforcement Learning}
\label{alg:grpo-objective}
\textbf{Input}: Trajectories $\{\tau_0^i\}_{i=1}^G$ sampled from current policy $\pi_\theta$; causal-oriented advantage $\hat{A}_{i,t}$; reference policy $\pi_{\text{ref}}$; hyperparameters $\beta$, $\epsilon$ \\
\textbf{Output}: Updated policy $\pi_\theta$
\begin{algorithmic}[1]
\FOR{each token $o^i_t$ in the sampled trajectories}
    \STATE Compute importance weight: $\rho_{i,t} \leftarrow \frac{\pi_\theta(o^i_t \mid s^i_t)}{\pi_{\text{old}}(o^i_t \mid s^i_t)}$
    \STATE Apply clipped weighting: $\Psi(\hat{A}_{i,t}) \leftarrow \text{clip}(\rho_{i,t}, 1 - \epsilon, 1 + \epsilon) \cdot \hat{A}_{i,t}$
\ENDFOR
\STATE Compute KL penalty: $\mu(\pi_\theta) \leftarrow D_{\mathrm{KL}}(\pi_\theta \,\|\, \pi_{\text{ref}})$
\STATE Compute joint objective $\mathcal{J}(\theta)$ as in Eq. (\ref{eq:grpo_objective})
\STATE Update policy $\pi_\theta$ by maximizing $\mathcal{J}_{\rm COPO}(\theta)$
\end{algorithmic}
\end{algorithm}

\section{More Discussion}
\label{sec_app:discussion}

\subsection{More Details for Motivation Experiment}
\label{sec_app:motivation_detail}
To probe how such hallucinations emerge, we conduct a controlled toy experiment that contrasts a MLLM with a text-only LLM under the same GRPO optimization settings and inspects their gradient patterns with respect to the input. 

\textbf{Setup.} We adopt GRPO post-training for both models to keep the optimization identical. The MLLM is LLaVA \cite{liu2023visual}, which encodes the image with a ViT, projects the image features into the same embedding space as text, and feeds the concatenated visual–text embeddings into a LLaMA decoder. The LLM is LLaMA \cite{grattafiori2024llama3herdmodels}, a text-only Transformer that receives no pixels. We choose this pair because (i) they share the same decoder backbone and comparable generation pipeline, (ii) the only substantive difference is the presence or absence of visual input, which lets gradient contrasts be attributed to the visual pathway, and (iii) both are strong, widely used baselines with public checkpoints. For each example, LLaVA takes the image-text pair (e.g., ``Is there a traffic signal light directing traffic?''), and LLaMA takes the same question plus a faithful textual description of the visible content. Both produce short factual answers. For analysis, we visualize their gradient distributions. We back-propagate to the visual input to obtain gradient saliency maps on LLaVA. And, we back-propagate to the token embeddings to gradient saliency on LLaMA. All maps are channel-aggregated and min–max normalized for display.

\textbf{Observation and analysis.} \textbf{Figure \ref{fig:motivation}} and \textbf{Figure \ref{fig:motivation_ex}} shows that gradient patterns differ markedly across models regardless of answer correctness. After GRPO post-training, the MLLM exhibits broader saliency that covers both task-relevant regions and visually salient but irrelevant background, whereas the LLM’s gradient saliency concentrates on semantic elements directly tied to the question (e.g., ``white bird'', ``traffic light''). Under outcome-based rewards, trajectories that yield correct answers while partially relying on background regularities are still positively reinforced, which may strengthen associations between background features and outcomes over time. Consequently, the MLLM may enhance the dependence on the background cues, which may cause hallucinations.

\subsection{Intuitive Causal Analysis}
\label{sec_app:intuitive_causal}
\begin{figure*}[t]
    \centering
    \begin{subfigure}[t]{0.49\textwidth}
        \centering
        \includegraphics[width=\textwidth]{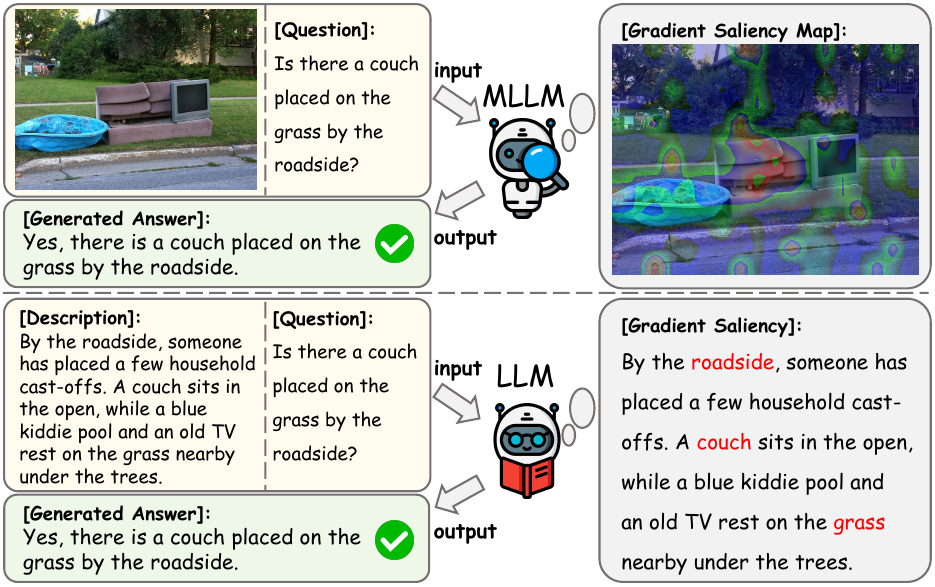}
        \caption{Scenario 1}
        \label{fig:motivation_ex1}
    \end{subfigure}
    \hfill
    \begin{subfigure}[t]{0.49\textwidth}
        \centering
        \includegraphics[width=\textwidth]{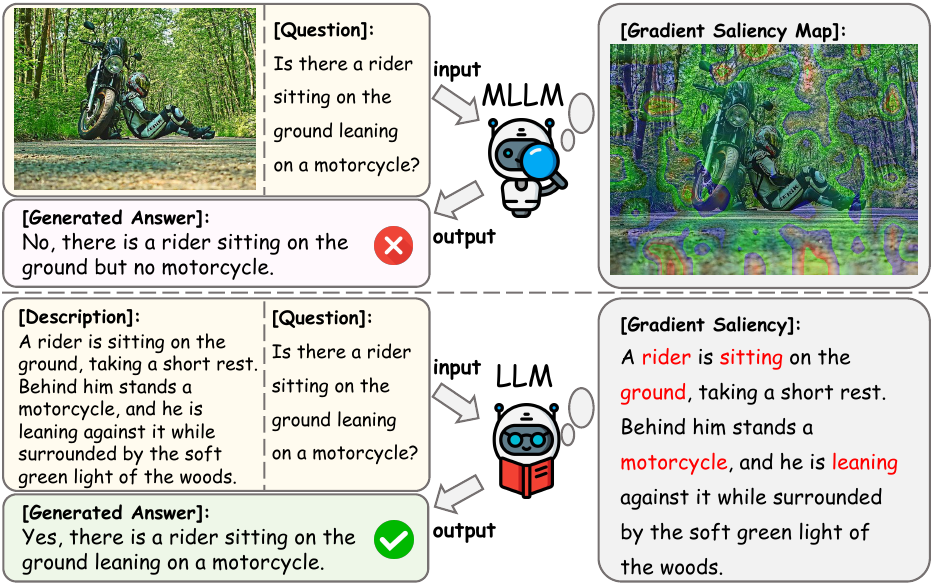}
        \caption{Scenario 2}
        \label{fig:motivation_ex2}
    \end{subfigure}
    \hfill
    \begin{subfigure}[t]{0.49\textwidth}
        \centering
        \includegraphics[width=\textwidth]{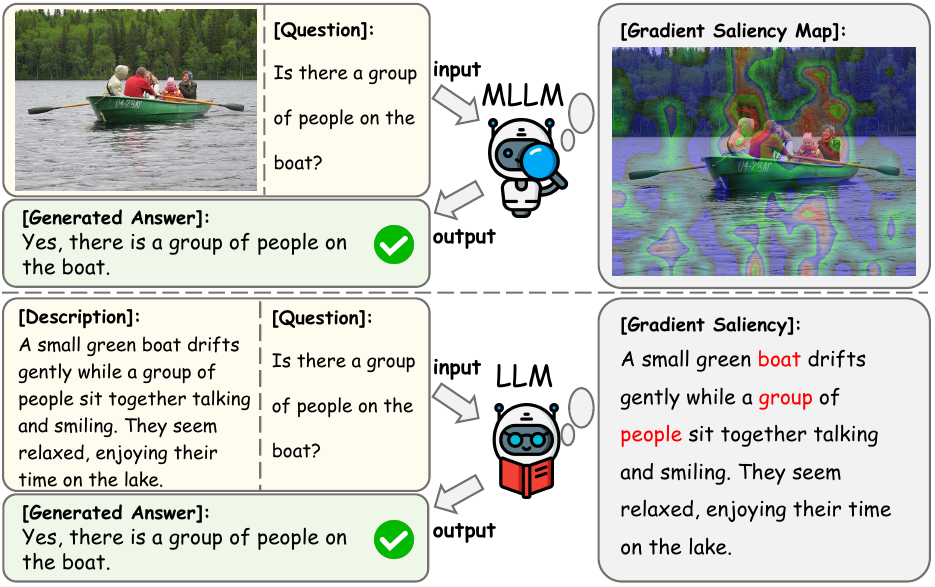}
        \caption{Scenario 3}
        \label{fig:motivation_ex3}
    \end{subfigure}
    \hfill
    \begin{subfigure}[t]{0.49\textwidth}
        \centering
        \includegraphics[width=\textwidth]{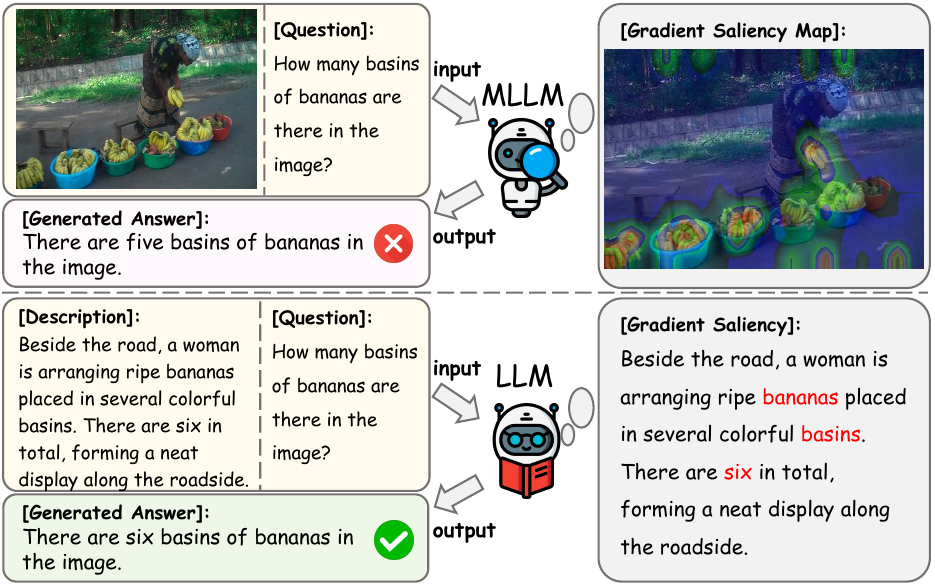}
        \caption{Scenario 4}
        \label{fig:motivation_ex4}
    \end{subfigure}
    \caption{More motivating results. Both MLLM and LLM are trained via GRPO. }
    \label{fig:motivation_ex}
\end{figure*}

To intuitively understand spurious correlations in MLLMs, we analyze concrete examples from \textbf{Figure \ref{fig:motivation}} and \textbf{Figure \ref{fig:motivation_ex}} through the lens of the SCMs introduced earlier. In our SCM, the ground-truth answer $Y$ (e.g., ``Yes, there is a white bird standing on the rock'') is determined by a set of causally relevant factors $L_c$, such as the presence of the bird, its color, and its position on the rock. These causal factors directly contribute to the accurate generation of the correct answer.

However, as demonstrated in \textbf{Figure \ref{fig:motivation}} and \textbf{Figure \ref{fig:motivation_ex}}, we observe that MLLMs often exhibit significant gradient saliency over irrelevant background features $L_s$, in addition to the task-relevant foreground features. For example, in \textbf{Figure \ref{fig:motivation_ex1}}, when asked ``Is there a couch placed on the grass by the roadside?'', the MLLM generates the correct answer ``Yes, there is a couch placed on the grass by the roadside''. However, the gradient saliency map shows that the model allocates considerable gradient saliency to irrelevant background regions such as the kiddle pool, despite the task-relevant foreground containing the couch.

Similarly, in \textbf{Figure~\ref{fig:motivation_ex2}}, the question ``Is there a rider sitting on the ground leaning on a motorcycle?'' results in the correct answer ``Yes, there is a rider sitting on the ground leaning on a motorcycle''. Yet, the model’s gradient saliency map reveals that the model is still heavily influenced by background regions, such as the surrounding trees, instead of focusing purely on the task-relevant object, the rider and motorcycle. 

Further examples, such as \textbf{Figure \ref{fig:motivation_ex3}} and \textbf{Figure \ref{fig:motivation_ex4}}, also reveal similar patterns. In \textbf{Figure \ref{fig:motivation_ex3}}, the MLLM answers correctly about the presence of a group of people on a boat, but the gradient saliency map shows unnecessary focus on background water areas rather than the people. In \textbf{Figure \ref{fig:motivation_ex4}}, the MLLM does not answers correctly about the number of banana basins, its gradient saliency map suggests undue influence from irrelevant background objects. These observations across multiple scenarios consistently demonstrate that MLLMs may rely on spurious correlations, where irrelevant background features unduly influence the model's reasoning process. 

These findings emphasize that the presence of spurious correlations in MLLMs, as captured by gradient saliency maps, may play a critical role in generating hallucinations. The model may rely on irrelevant background cues, even when the final answer may be correct. It is necessary to guide MLLMs to focus on relevant information and reduce the influence of background elements, ultimately reducing the likelihood of hallucinations.

\subsection{Comparison and Uniqueness Discussion}
\label{sec_app:comparison_uniqueness}
``Causal sufficiency and necessity'' is a foundational concept in causal theory, formally introduced in the book ``Causality'' \cite{pearl2009causality}. Recent works have explored this concept in various contexts, including CoT-based reasoning in LLMs \cite{yu2025causal}. To clarify the scope and positioning of our framework and distinguish it from existing studies, we summarize the key differences in modeling assumptions and optimization objectives.

\subsubsection{Different Motivations and Core Ideas.}
While both methods are inspired by causal theory, our framework is motivated by hallucinations in MLLMs, which are often caused by spurious correlations between task-irrelevant background cues and the final answer. We formulate spurious correlations from a causal perspective and introduce token-level constraints based on causal sufficiency and necessity to guide generation. In contrast, \cite{yu2025causal} aim to improve interpretability in LLM reasoning, focusing on analyzing the contribution of intermediate reasoning steps. \cite{yu2025causal} leverage post-hoc attribution to assess whether specific tokens are causally linked to the model's final decision.

\subsubsection{Different Methodologies and Experiments.}
Our method integrates causal constraints into the optimization through reinforcement learning. We construct a causal completeness reward that encourages the model to consider both sufficiency and necessity tokens for correct answers. This allows the model to gradually shift attention toward more causally relevant information during optimization. In comparison, \cite{yu2025causal} adopt a causal probing strategy, evaluating fixed model outputs without modifying the learning procedure. \cite{yu2025causal} propose a causal attribution method intended to interpret model predictions without altering the training procedure. In experiments, our framework is assessed on multiple MLLM benchmarks with hallucination metrics to examine its practical effectiveness in reducing hallucinations. On the other hand, \cite{yu2025causal} focus on qualitative and interpretability-driven analysis in text-only settings.

In summary, these differences reflect distinct goals and methodological choices. Our approach targets hallucination mitigation in multimodal tasks through reward-based optimization, whereas \cite{yu2025causal} pursue causal interpretability for language reasoning.

\begin{figure*}[t]
    \centering
    \centering
    \includegraphics[width=\textwidth]{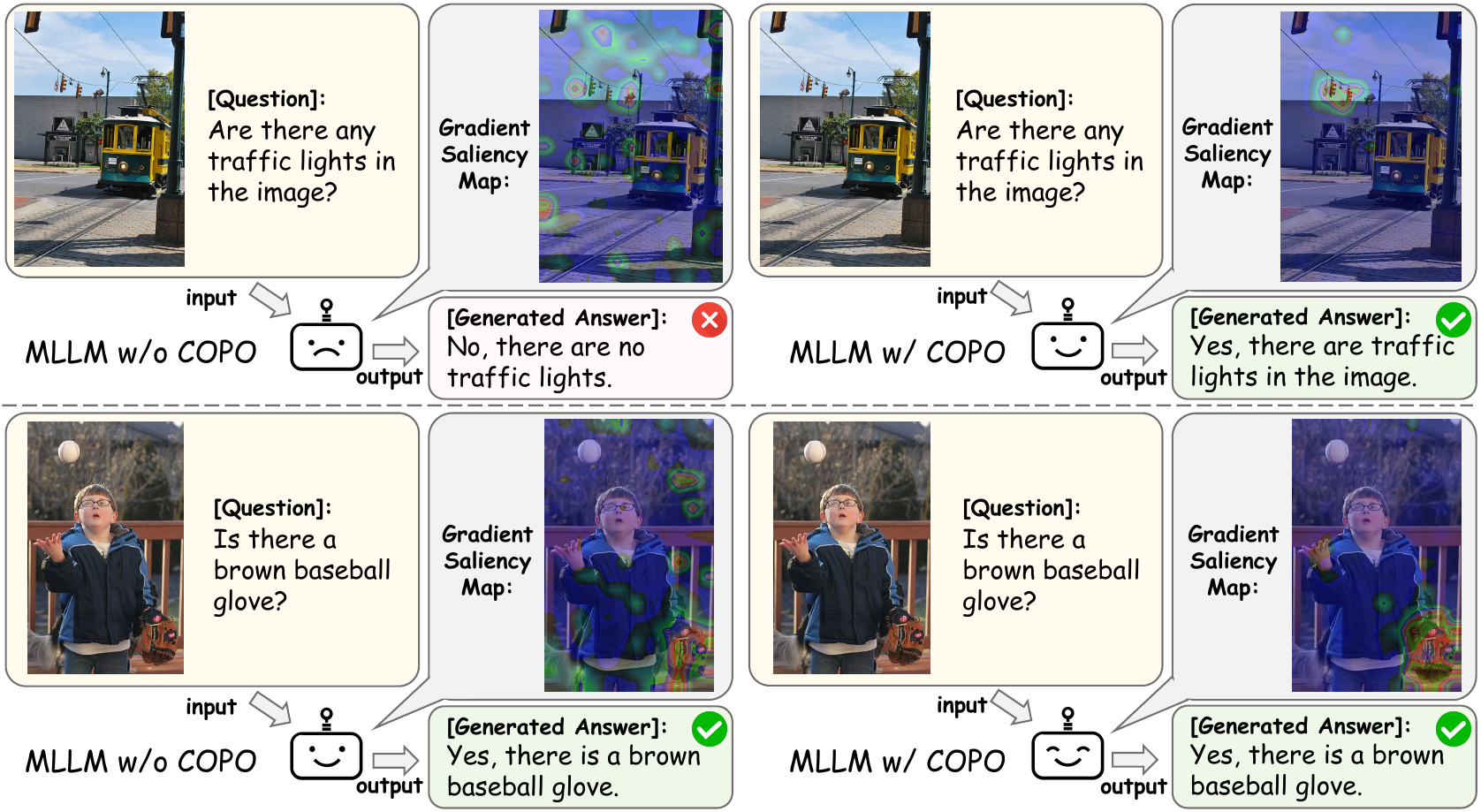}
    \caption{More illustration of causal completeness reward effect.}
    \label{fig:causal_effect_ex}
\end{figure*}

\subsection{Intuitive Causal Completeness Reward Effect}
\label{sec_app:causal_reward_effect}
In this section, we provide a more intuitive explanation of the effect of the causal completeness reward on mitigating hallucinations in MLLMs.

Hallucinations in MLLMs often arise when the model generates plausible-sounding responses that are not grounded in the visual evidence, typically due to reliance on non-causal background information. \textbf{Figure \ref{fig:exp_compare}} and \textbf{\ref{fig:causal_effect_ex}} illustrate this effect by comparing the behavior of an MLLM with and without our proposed COPO. Without COPO, the gradient saliency of the model is typically spread across both task-relevant regions and irrelevant background features. For example, when asked about the presence of traffic lights in \textbf{Figure \ref{fig:causal_effect_ex}}, the MLLM might generate the incorrect response ``No, there are no traffic lights'', despite background elements like building façades influencing its decision. After applying COPO, the model shifts its attention to task-relevant regions, such as the traffic light itself, producing the correct response ``Yes, there are traffic lights in the image''. 

Additionally, in cases where the MLLM initially provides the correct answer, COPO can further refine the model’s reasoning process. For example, in \textbf{Figure~\ref{fig:motivation}}, the model answers correctly, ``Yes, there is a brown baseball glove'', but without COPO, the gradient saliency is distributed over both the glove and background regions. After applying COPO, the gradient saliency becomes significantly more concentrated on the foreground object (the glove), reinforcing the model's focus on causally relevant features, which may improve its reasoning process.

These results suggest that COPO can make the reasoning of the model more consistent with the causal relevant elements, thus minimizing the influence of irrelevant background information and reducing the likelihood of hallucinations.

\subsection{Qualitative Analysis}
\label{sec_app:qualitative_analysis}
\begin{figure*}
    \centering
    \includegraphics[width=0.9\textwidth]{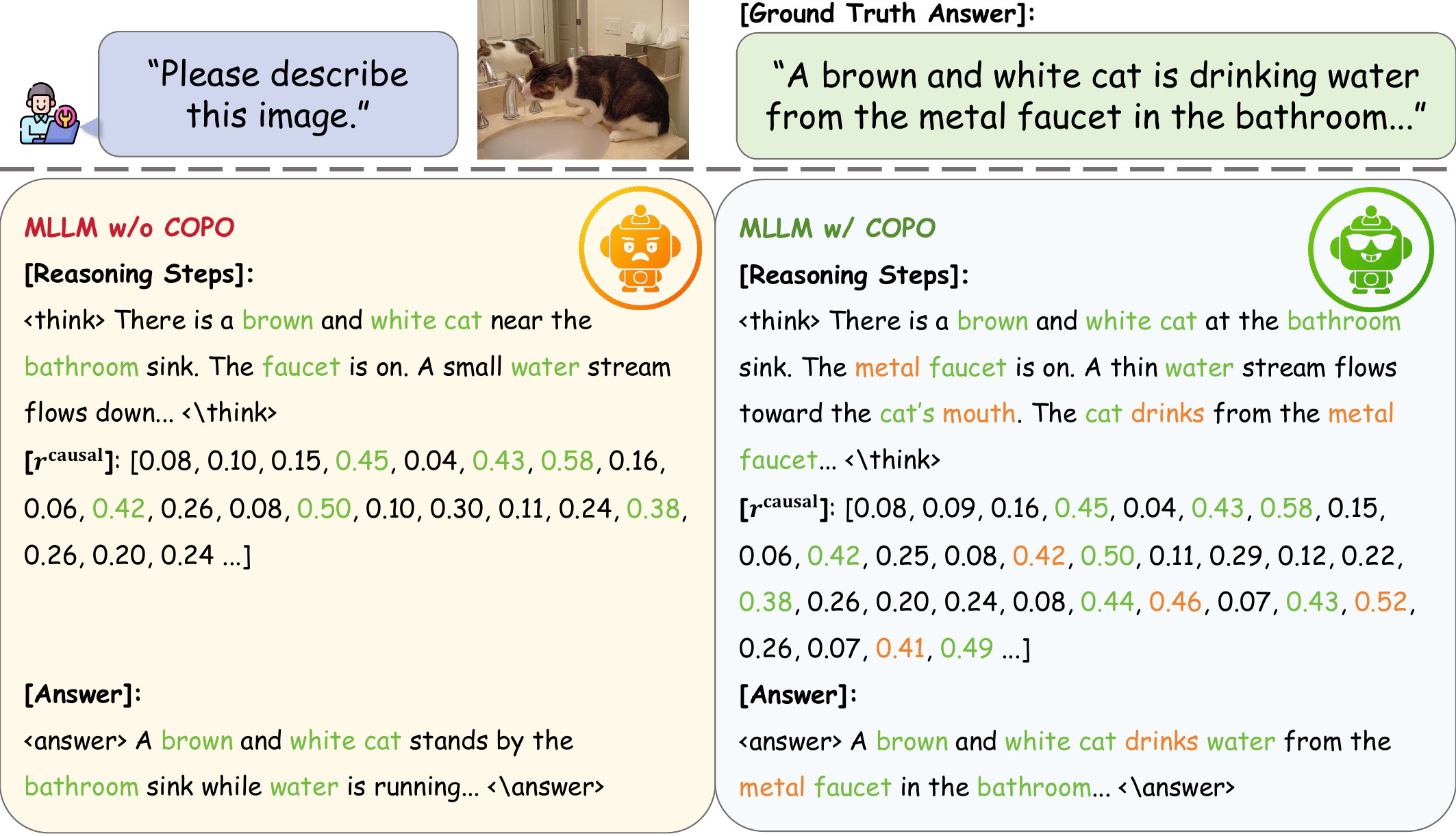}
    \caption{Qualitative analysis of causal completeness reward. 
    }
    \label{fig:qualitative_analysis}
\end{figure*}

To better illustrate the behavior of our causal completeness reward, we conduct a qualitative analysis on an example from the image captioning task. 
As shown in \textbf{Figure \ref{fig:qualitative_analysis}}, the input image depicts a brown and white cat drinking water from a metal faucet in a bathroom sink. 

The MLLM w/o COPO identifies several relevant visual attributes such as brown, white, cat, bathroom, sink, and water. These content words obtain moderately high causal rewards, while many tokens associated with the actual action (e.g., drinking, mouth, faucet interaction) are absent from the reasoning steps. As a consequence, the trajectory contains incomplete causal evidence, and its causal reward remains relatively low. The final answer, though partially correct, fails to state that the cat is drinking from the faucet.

For the MLLM w/ COPO, the reasoning steps become more causally aligned with the true visual factors. Tokens capturing critical relations-—such as metal faucet, thin water stream, cat’s mouth, and drinks—-now appear explicitly and receive substantially higher causal rewards. Meanwhile, non-informative connectives continue to obtain low scores, showing that the causal reward selectively emphasizes semantically grounded, evidence-bearing tokens. This produces a reasoning trajectory with a higher overall causal completeness.

These results imply that COPO can reliably elevate the causal completeness reward of tokens that genuinely contribute to a correct description, and sequences with higher causal reward correspond to significantly more accurate, evidence-grounded outputs. This demonstrates the effectiveness of COPO.

\subsection{Broader Impacts and Limitations}
\label{sec_app:impact_limitation}
In this subsection, we briefly illustrate the broader impacts and limitations of this work.

\paragraph{Broader Impacts. }
This work investigates hallucination mitigation in MLLMs and proposes a causal-oriented policy optimization (COPO) framework. It aims to reduce hallucinations by encouraging outputs that are both causally sufficient and necessary for correct answers. Our approach offers a conceptual bridge between causal inference and hallucinations of MLLMs, potentially inspiring further research into causal-aware generation strategies and intervention-based learning for foundation models. Extensive experiments across diverse benchmarks validate the effectiveness of our COPO in mitigating hallucinations.

\paragraph{Limitations. }
Our implementation is based on open-source MLLMs and LLMs, and we have not evaluated models at the scale of 72B or above 100B parameters. Future work may explore applying COPO to larger-scale models to further assess its robustness and scalability.

\section{Datasets}
\label{sec_app:dataset}
In this section, we provide a brief overview of the datasets used in our experiments. We utilize publicly available datasets. 

\begin{itemize}
    \item MSCOCO \cite{lin2014microsoft} is a large-scale benchmark designed for image captioning, object detection, and scene understanding. The 2014 version contains 82,783 training images and 40,504 validation images, each annotated with five human-written captions describing the image content. The dataset features diverse everyday scenes with multiple objects in context, making it suitable for evaluating visual grounding, object recognition, and multimodal generation tasks. Its high-quality annotations and broad coverage of object categories have made it a standard resource in vision-language research.
    \item $V^*$ dataset \cite{wu2024v} is constructed based on COCO2017, specifically tailored for evaluating visual search and fine-grained grounding capabilities in multimodal models. It features high-resolution natural images where the target objects are often small, densely distributed, and surrounded by clutter. This setting poses a significant challenge for MLLMs, as models must detect and localize fine-scale objects while integrating visual and linguistic cues. The dataset is particularly useful for assessing models' abilities in precise visual grounding and attention allocation within complex scenes. 
    \item ArxivQA \cite{li2024multimodal} is a multimodal question-answering dataset that focuses on scientific diagrams and structured visual representations extracted from research papers on arXiv. The dataset consists of questions grounded in visual elements such as line charts, bar graphs, tables, and flow diagrams. Answering these questions requires cross-modal understanding of the visual layout, semantic parsing of textual annotations, and precise diagram interpretation. This subset effectively evaluates the abilities of MLLMs to align textual queries with structured visual information and reason over schematic content.
    \item ThinkLite-VL \cite{wang2025sota} is designed to benchmark complex multimodal reasoning abilities. It includes diverse question-answering tasks that span several categories such as mathematical reasoning, physical and commonsense inference, temporal understanding, and causal analysis. Each question is paired with multimodal context (image and/or text) that requires step-by-step reasoning rather than shallow pattern recognition. The dataset emphasizes robustness and generalization in reasoning, making it particularly suitable for evaluating models' inference chains and causal understanding in multimodal scenarios. 
\end{itemize}

\section{Baselines}
\label{sec_app:baseline}
In this section, we provide a brief overview of the baselines used in our experiments.

\begin{itemize}
    \item DoLa \cite{chuang2023dola} enhances factual grounding in MLLMs by leveraging the contrast between shallow and deep semantic features across the model layers. During decoding, it compares early-layer and late-layer activations to identify inconsistencies and promote factually grounded token selection. This mechanism helps suppress hallucinations by ensuring that surface-level visual signals are supported by deeper semantic representations. 
    \item OPERA \cite{huang2024opera} mitigates hallucinations by identifying and calibrating over-trusted visual tokens. It analyzes self- and cross-attention distributions to locate tokens that disproportionately dominate the model's generation process. By down-weighting such tokens through attention re-scaling, OPERA reduces hallucinated content and improves factual alignment in generated text.
    \item VCD \cite{leng2024mitigating} improves visual grounding by enforcing contrastive consistency between relevant and irrelevant visual-text pairs at the representation level. It introduces a contrastive training objective that pulls matched image-text features closer while pushing mismatched ones apart. This contrastive signal enhances the model's ability to focus on visually grounded elements and reduces reliance on hallucinated concepts.
    \item GPT-4o \cite{achiam2023gpt} is OpenAI's flagship MLLM, designed to handle both textual and visual inputs with high fidelity. It incorporates proprietary training strategies and massive-scale data to achieve strong cross-modal alignment, reasoning, and generation capabilities. GPT-4o represents one of the best-performing commercial MLLMs, making it a valuable reference point for evaluating model performance on real-world multimodal tasks, including hallucination robustness. 
    \item Qwen-VL-Max \cite{bai2023qwenvlversatilevisionlanguagemodel} is a high-capacity vision-language model developed by Alibaba, designed for detailed visual understanding and instruction following. It extends the Qwen-VL family by incorporating larger backbone architectures and fine-grained alignment techniques, supporting multi-turn interaction, high-resolution image inputs, and extended reasoning abilities. Qwen-VL-Max is trained on a broad set of image-text pairs and instruction-following data, and optimized for tasks such as image captioning, VQA, and OCR. Due to its closed-source nature, we use its official public API for inference-based comparison in our evaluation.
    \item InternVL-1.5 \cite{chen2024far} an open-source large vision-language model introduced by Shanghai AI Laboratory. It builds upon previous InternVL versions, incorporating enhanced multi-modal alignment, improved image understanding, and more efficient training techniques. The model excels in zero-shot and few-shot settings, making it a strong benchmark for open-source multimodal research. 
    \item LLaVA-OneVision \cite{li2024llava} is an advanced open-source vision-language model that unifies training across diverse image resolutions and visual contexts. It improves upon previous LLaVA models by introducing resolution-aware learning and high-resolution instruction tuning, enabling robust performance on complex visual tasks. As a strong open-source baseline, LLaVA-OneVision provides a reliable comparison point for models targeting high-resolution and fine-grained visual understanding. 
    \item Qwen2.5-VL \cite{bai2025qwen2} is a powerful multimodal extension of the Qwen language model series, developed by Alibaba. It incorporates vision encoders aligned with large language models and supports instruction tuning for a wide range of vision-language tasks. With both 7B and 32B variants, Qwen2.5-VL demonstrates strong performance in image captioning, visual QA, and reasoning, serving as a representative of scalable, open-source MLLMs. 
    \item DeepEyes \cite{zheng2025deepeyes} is a vision-language model that emphasizes external tool usage and modular vision-language reasoning. It integrates structured reasoning modules and supports external search or computation for improved factuality and interpretability. DeepEyes has shown effectiveness in reducing hallucinations through hybrid symbolic-neural workflows, making it a particularly relevant baseline for hallucination mitigation studies. 
    \item Grounding DINO \cite{liu2024groundingdinomarryingdino} is a multi-stage system that integrates object detection and open-set visual grounding within a single framework. By combining powerful transformer backbones with structured reasoning workflows, Grounding DINO excels at referring expression comprehension and open-vocabulary detection, offering precise and interpretable grounding results in complex scenes.
    \item SEAL \cite{wu2024v} is a workflow-based system designed for fine-grained visual understanding, especially in high-resolution visual search tasks. It incorporates structured pipelines including proposal generation, visual grounding, and language reasoning to address small-object detection and disambiguation. SEAL's multi-stage reasoning capabilities offer high accuracy at the cost of complexity, and it serves as a strong non-end-to-end baseline.
    \item DyFo \cite{li2025dyfo} introduces a dynamic fusion framework that adaptively integrates visual and textual features through modular fusion layers and reasoning controllers. It leverages intermediate symbolic representations to guide multimodal inference, enhancing interpretability and performance on challenging tasks. DyFo represents a hybrid architecture combining structured reasoning and deep learning, useful for assessing generalization under complex scenarios. 
    \item ZoomEye \cite{shen2024zoomeye} is a multi-stage system that targets high-resolution image analysis by progressively zooming in on relevant image regions. It mimics human attention by narrowing focus to informative areas before performing visual-language reasoning, significantly improving performance on cluttered and detail-rich images. ZoomEye's pipeline exemplifies how structured workflows can improve precision and reduce hallucinations in vision-language tasks. 
    \item DeCo \cite{wang2025mllmseedynamiccorrection} is a training-free, model-agnostic decoding framework proposed to mitigate hallucinations in MLLMs. The core insight of DeCo is that MLLMs often correctly recognize visual objects in their preceding (intermediate) layers, but this information is suppressed in deeper layers due to strong language model priors, resulting in hallucinated outputs. DeCo dynamically selects the most informative preceding (anchor) layer, and proportionally integrates its knowledge into the final layer's output logits during inference. This adjustment recalibrates the predicted token probabilities, enhancing the accuracy of object descriptions while reducing hallucination rates. DeCo is compatible with various decoding strategies—such as greedy search, nucleus sampling, and beam search—and can be seamlessly applied to different MLLMs without additional training. Extensive experiments show that DeCo significantly reduces hallucinations on image captioning and visual question answering benchmarks, with only a modest increase in inference latency compared to baseline methods.
    \item VTI \cite{liu2025reducing} introduces a method called Visual and Textual Intervention (VTI), designed to reduce hallucinations in large vision-language models (LVLMs). VTI works by steering the latent space representations of both the vision encoder and text decoder during inference. This test-time intervention enhances the stability of vision features and improves the alignment between vision and text, without the need for additional training. The technique involves pre-computing the directions of feature shifts in the latent space based on a set of training examples. These directions are then applied consistently to all queries during inference. The method is task- and dataset-agnostic, making it widely applicable without incurring additional computational costs. VTI has shown significant effectiveness in reducing hallucinations across multiple evaluation benchmarks.
    \item HA-DPO \cite{zhao2023beyond} is a fine-tuning method designed to reduce hallucinations in Large Vision-Language Models (LVLMs). The key idea is to treat hallucination mitigation as a preference learning problem: for each image, the model is provided with a pair of responses—one faithful and one hallucinated—and DPO is used to directly optimize the model to prefer the faithful response. To construct training data, the authors curate paired hallucination/non-hallucination outputs with consistent style, and apply lightweight LoRA fine-tuning on multiple LVLM architectures such as MiniGPT-4, InstructBLIP, and LLaVA-1.5. Experiments on standard hallucination benchmarks (e.g., POPE, SHR) show that HA-DPO effectively suppresses visual hallucinations and improves response reliability.
    \item CLIP (OHD) \cite{liu2024investigating} presents a focused analysis of object-hallucination phenomena in the commonly used visual encoder CLIP, which serves as the backbone for many large vision-language models (LVLMs). The authors first introduce a dedicated benchmark called OHD-Caps (Object Hallucination Detection-Caps) comprising paired positive captions and counterfactual negatives (captions with nonexistent or removed objects) drawn from datasets such as COCO, Flickr30K and NoCaps. Their empirical findings show that even the CLIP encoder in isolation is prone to mistaking hallucinated objects, indicating that hallucination is not purely a vision-language fusion issue. To mitigate this, they apply a counterfactual data-augmentation strategy to fine-tune the CLIP encoder using a fine-grained object-level contrastive loss on the OHD-Caps dataset, showing large reductions in object-hallucination rates and improvements when the tuned encoder replaces the vanilla CLIP in downstream LVLMs.
    \item POVID \cite{zhou2024aligning} presents a novel approach for reducing hallucination in vision-language models (VLLMs) by explicitly aligning the visual and language modalities through preference‐based fine-tuning. The authors propose the method named POVID, which generates preference pairs by first using a large multimodal model (e.g., GPT-4V) to inject plausible hallucinations into correct responses, then distorting input images to trigger inherent hallucination behaviours of the VLLM. These preference pairs (faithful vs hallucinated) are then used to fine-tune the model via Direct Preference Optimization (DPO). Experimental results across multiple image-instruction benchmarks show that this modality alignment strategy reduces hallucination error rates and improves overall generation reliability.
    \item CSR \cite{zhou2024calibrated} introduces a novel mechanism for improving the alignment of visual and language modalities in large vision-language models (LVLMs) by using self-rewarding fine-tuning. Specifically, the authors propose to generate model-internal reward signals calibrated to identify hallucination versus faithful outputs, and then use these self-derived rewards to fine-tune the model without reliance on extensive human annotation. Through experiments on image‐instruction benchmarks, the paper demonstrates that the self-rewarding framework reduces hallucination rates and enhances output fidelity while maintaining inference efficiency.
    \item GCPO \cite{gu2025group} introduces a novel RL-style fine-tuning method designed for large-scale language models that addresses the limitations of existing group-wise preference learning approaches. The key innovation is to integrate causal structure among candidate responses by (1) applying a causally informed reward adjustment that accounts for interactions (e.g., complementarity or contradiction) among grouped candidates; and (2) incorporating a KL-regularization term aligning the policy with a causally projected reference distribution. Extensive experiments on multiple reasoning benchmarks demonstrate that GCPO consistently outperforms prior methods such as Group Relative Policy Optimization (GRPO) by achieving better calibration and reward alignment.
\end{itemize}

\section{Benchmarks}
\label{sec_app:benchmark}
In this section, we provide a brief overview of the benchmarks used in our experiments.

\begin{itemize}
    \item CHAIR \cite{rohrbach2018object} is a standard metric for measuring object hallucination in image captioning. It compares model-generated captions against ground-truth object annotations from MSCOCO, quantifying hallucination at two levels: sentence-level (CHAIR$_S$), which computes the fraction of captions containing hallucinated objects, and instance-level (CHAIR$_I$), which measures the fraction of hallucinated object mentions among all mentioned objects. Following previous works~\cite{huang2024opera}, we use 500 randomly sampled images from the MSCOCO 2014 validation set and adopt the fixed prompt ``Please help me describe the image in detail.''
    \item Perception-Oriented Perturbation Evaluation (POPE) \cite{li2023evaluating} introduces controlled adversarial and semantic perturbations into visual inputs to assess hallucination sensitivity and robustness. It contains three subsets—Adversarial, Popular, and Random—that evaluate whether a model can avoid producing hallucinated content when facing visually misleading or out-of-distribution inputs. POPE is a standard benchmark for hallucination detection and mitigation evaluation. 
    \item MME \cite{fu2024mmecomprehensiveevaluationbenchmark} is a diagnostic benchmark focusing on fine-grained multi-modal capabilities, particularly in assessing text generation quality across sub-skills such as attribute grounding, object counting, relation understanding, OCR, and commonsense reasoning. For each model, we extract the averaged score across all text-related subskills. Evaluation is based on matching predicted answers against ground-truth labels using the official toolkit.
    \item GPT-4o Assistance \cite{achiam2023gpt} is an open-ended caption assessment benchmark designed to evaluate model performance in factuality, coherence, and informativeness of generated descriptions. It uses GPT-4o to compare captions from different models on a set of 100 randomly selected images from the MSCOCO 2014 validation set. For each image, two model outputs are compared side-by-side, and GPT-4o is prompted to assess them along three axes: Accuracy (A), measuring the factual correctness of the description; Correctness (C), assessing logical consistency; and Detailedness (D), capturing the richness of content. This evaluation follows the same format as~\cite{huang2024opera, leng2024mitigating} using a standardized prompt template. 
    \item $V^*$ Bench \cite{wu2024v} is a high-resolution benchmark built on the COCO2017 dataset, specifically curated for fine-grained object attribute and spatial relation evaluation. It contains complex scenes with small, dense visual targets and requires precise object localization and captioning under high-resolution settings. This benchmark is designed to test a model's ability to maintain grounding accuracy and attribute fidelity in visually cluttered scenarios. 
    \item HR-Bench(4K/8K) \cite{wang2025divide} provides a series of vision-language evaluation tasks based on ultra high-resolution images at 4K and 8K scales. The benchmark includes factual QA, reasoning, and captioning tasks over visually complex environments, often involving minute object details. It challenges models to maintain accurate semantic understanding under extreme resolution and compositional complexity, making it ideal for evaluating resolution scalability and hallucination resistance.
    \item refCOCO \cite{caesar2018coco} is a benchmark designed for evaluating object grounding via referring expressions in natural images. Each instance provides an image and a natural language expression referring to a specific object, and the task is to localize the correct object. The expressions in refCOCO often include both appearance and spatial relations, making it a strong testbed for assessing multimodal models' grounding fidelity and cross-modal understanding under relatively simple sentence structures. 
    \item refCOCO+ \cite{caesar2018coco} is a variant of refCOCO that restricts referring expressions to exclude spatial terms. This forces models to rely more heavily on visual attributes—such as color, size, and object type—rather than relative positioning. As a result, refCOCO+ is particularly useful for evaluating a model's capacity to attend to fine-grained visual details and object characteristics when grounding textual descriptions. 
    \item refCOCOg \cite{kazemzadeh2014referitgame} extends the previous benchmarks by providing longer and more descriptive referring expressions that often include multiple attributes and complex sentence structures. Unlike refCOCO and refCOCO+, it contains full-image annotations rather than cropped regions, increasing contextual ambiguity. This makes refCOCOg a challenging benchmark for assessing how well multimodal models can handle long-range dependencies and integrate multiple pieces of information for accurate grounding. 
    \item ReasonSeg \cite{lai2024lisa} is a recent benchmark that evaluates visual segmentation tasks requiring multi-hop reasoning. Each instance includes an image, a question, and a target segmentation mask, requiring the model to understand complex dependencies between visual regions and linguistic cues. It is particularly useful for assessing how well a model can integrate reasoning with pixel-level grounding and visual comprehension.
    \item MathVista \cite{lu2023mathvista} is a multimodal benchmark composed of visual math problems that require understanding diagrams, plots, or geometric figures. It tests spatial reasoning and diagram interpretation in mathematical contexts, making it particularly suitable for evaluating visual grounding in structured mathematical domains. MathVista highlights the challenge of aligning symbolic and visual information for accurate problem-solving. 
    \item MathVerse \cite{zhang2024mathverse} provides a broad spectrum of math-related visual-language problems ranging from elementary arithmetic to higher-order logic. It combines equation understanding with diagram-based inputs, evaluating the model's ability to perform compositional reasoning across modalities. MathVerse is a robust benchmark for assessing general-purpose multimodal mathematical reasoning.
    \item MathVision \cite{wang2024measuring} focuses on assessing visual mathematical reasoning through tasks such as geometry, bar charts, and algebraic diagram interpretation. It contains highly structured problems where success depends on identifying visual evidence and integrating it with symbolic operations. MathVision is particularly valuable for evaluating reasoning robustness and multi-hop inference over structured inputs.
    \item WeMath \cite{qiao2024we} introduces a diverse set of visual math challenges involving open-ended questions that mimic real-world math education scenarios. The tasks test a model's capacity for high-level reasoning, multi-step solution planning, and causal inference based on visual stimuli. WeMath promotes evaluation under pedagogical constraints and human-like reasoning processes.
    \item DynaMath \cite{zou2024dynamath} emphasizes dynamic visual reasoning over animated or multi-frame visual content. It includes motion-based problems that require temporal reasoning, causal tracking, and spatial transformation understanding. DynaMath serves as a comprehensive benchmark for temporal multimodal reasoning beyond static-image settings.
    \item LogicVista \cite{xiao2024logicvista} targets visual logic reasoning through a collection of puzzles and structured visual problems. It focuses on logic rule inference, pattern completion, and counterfactual reasoning, challenging models to deduce correct answers through symbolic and visual clues. LogicVista evaluates higher-order causal and logical reasoning in visually constrained contexts.
\end{itemize}

\section{Implementation Details}
\label{sec_app:implementation}
Our implementation builds upon open-source reinforcement learning frameworks, modified to incorporate a causal completeness reward for hallucination mitigation. We conduct training on four representative open-source vision-language models: InstructBLIP \cite{dai2023instructblipgeneralpurposevisionlanguagemodels}, MiniGPT-4 \cite{zhu2023minigpt}, LLaVA-1.5 \cite{liu2024improved}, and Qwen-VL \cite{bai2023qwenvlversatilevisionlanguagemodel}. All models are initialized from publicly released Hugging Face checkpoints. Training is conducted on A100 GPU clusters using FSDP with parameter and optimizer offloading enabled to ensure memory efficiency. We use vLLM for rollout generation, with GPU memory utilization capped at 60\% and support for long-sequence prefill enabled.

The training follows the GRPO setup, with each batch consisting of 512 sampled prompts and a mini-batch size of 256. We use a causal completeness reward function that balances causal sufficiency and necessity, with weights $\lambda_s = 0.35$ and $\lambda_n = 0.35$. The maximum prompt length is set to 1,024, and the maximum response length is extended to 8,192 tokens to support complex reasoning. The KL divergence regularization is enabled with a low-variance KL loss, and the reference model shares the same architecture as the actor. Learning rate is fixed at $1e^{-6}$. COPO is applied in an offline manner. All rollouts used for computing causal rewards are pre-collected, and the optimization operates entirely over these fixed trajectories. This design avoids repeated online sampling during training and keeps the computational overhead within a reasonable and manageable range.
All experiments are executed with mixed precision. The main code is provided in supplementary materials.

For evaluation, we follow the exact protocol and data splits from prior works. For CHAIR~\cite{rohrbach2018object}, we use 500 images from the MSCOCO 2014 validation set with the fixed prompt ``Please help me describe the image in detail'', reporting both $\mathrm{CHAIR_S}$ and $\mathrm{CHAIR_I}$ metrics. For POPE~\cite{li2023evaluating}, we follow the VQA-style evaluation on 500 MSCOCO images, using six structured object-centric questions per image across three splits (random, popular, adversarial), and report averaged F1 scores. For text quality, we evaluate on MME \cite{fu2024mmecomprehensiveevaluationbenchmark}. MME results are based on the averaged scores of 12 text-related sub-skills. For GPT-4o assisted evaluation, we sample 100 images from the COCO validation set and adopt the annotation protocol introduced by~\citet{huang2024opera,leng2024mitigating}, where GPT-4o evaluates two model outputs per image on three axes: Accuracy (A), Correctness (C), and Detailedness (D). Both text quality and GPT-4o assisted evaluation are conducted base on LLaVA-1.5.

\begin{table*}[t]
    \centering
    \caption{Results on high-resolution visual understanding. The best results are highlighted in \textbf{bold}.}
    \label{tab:exp_hr}
    \setlength{\tabcolsep}{1mm}
    \begin{tabular}{l c | c c c|c c c|c c c } 
    \toprule
    \multirow{2}{*}{\textbf{Model}} & \multirow{2}{*}{\textbf{\makecell{Param \\ Size} }} &  \multicolumn{3}{c|}{$V^*$ Bench} & \multicolumn{3}{c|}{HR-Bench 4K} & \multicolumn{3}{c}{HR-Bench 8K}\\
    &  & \textbf{Attr} & \textbf{Spatial} & \textbf{Overall} & \textbf{FSP} & \textbf{FCP} & \textbf{Overall} & \textbf{FSP} & \textbf{FCP} & \textbf{Overall}\\
    \midrule
    GPT-4o~\cite{achiam2023gpt}   & - & - & - &66.0 & 70.0 & 48.0 &   59.0  & 62.0  & 49.0 & 55.5 \\
    Qwen-VL-max~\cite{bai2023qwenvlversatilevisionlanguagemodel}   & - & - & - &- & 65.0 & 52.0 & 58.5  & 54.0  & 51.0 & 52.5 \\
    \midrule
    SEAL~\cite{wu2024v}  & 7B & 74.8 & 76.3 & 75.4 & - & - & - & - & - & - \\
    DyFo~\cite{li2025dyfo}  & 7B & 80.0 & 82.9 & 81.2 & - & - & - & - & - & - \\
    ZoomEye~\cite{shen2024zoomeye}  & 7B & 93.9 & 85.5 & 90.6 & 84.3 & 55.0 & 69.6 & 88.5 & 50.0 & 69.3 \\
    \midrule
    InternVL-1.5~\cite{chen2024far}  & 26B  & - & - & - & 69.5 & 51.8 & 60.6 & 69.3 & 48.5 & 57.9 \\
    LLaVA-OneVision~\cite{li2024llava}  & 7B  & 75.7 & 75.0 & 75.4 & 72.0 & 54.0 & 63.0 & 67.3 & 52.3 & 59.8 \\
    Qwen2.5-VL~\cite{bai2025qwen2}  & 7B & 73.9 & 67.1 & 71.2 & 85.2 & 52.2 & 68.8 & 78.8 & 51.8 & 65.3 \\
    Qwen2.5-VL~\cite{bai2025qwen2}  & 32B & 87.8 & 88.1 & 87.9 & 89.8 & 58.0 & 73.9 & 84.5 & 56.3 & 70.4 \\
    DeepEyes \cite{zheng2025deepeyes} & 7B & 91.3 & 88.2 & 90.1 & 91.3 & 59.0 & 75.1 & 86.8 & 58.5 & 72.6 \\
    \midrule
    \textbf{Ours}  & 7B & \textbf{95.0} & \textbf{91.1} & \textbf{93.4} & \textbf{92.7} & \textbf{61.1} & \textbf{76.7} & \textbf{90.1} & \textbf{62.8} & \textbf{77.2} \\
    $\Delta$ (\textit{vs} SOTA) & - & +1.1 & +2.9 & +2.8 & +1.4 & +2.1 & +1.6 & +1.6 & +4.3 & +4.6 \\
    \bottomrule
    \end{tabular}
\end{table*}

\begin{table*}[t]
    \centering
    \caption{Performance on grounding fidelity. The best results are highlighted in \textbf{bold}.}
    \label{tab:exp_gfhs}
    \begin{tabular}{l c | c c c c  } 
    \toprule
    \multirow{2}{*}{\textbf{Model}} & \multirow{2}{*}{\textbf{\makecell{Param \\ Size} }} &  \multirow{2}{*}{\textbf{refCOCO}} &  \multirow{2}{*}{\textbf{refCOCO+}} &  \multirow{2}{*}{\textbf{refCOCOg}} &  \multirow{2}{*}{\textbf{ReasonSeg}} \\
    & & & & &  \\
    \midrule
    Grounding DINO~\cite{liu2024groundingdinomarryingdino} & 7B  & 88.2 & 75.9 & 87.0 \\
    Qwen2.5-VL~\cite{bai2025qwen2} & 7B & 89.1 & 82.6 & 86.1 & 68.3  \\
    DeepEyes \cite{zheng2025deepeyes}& 7B & 89.8 & 83.6 & 86.7 & 68.6  \\
    \midrule
    \textbf{Ours} & 7B & \textbf{91.2} & \textbf{85.6} & \textbf{88.0} & \textbf{70.2} \\
    $\Delta$ (\textit{vs}SOTA) & - & +1.4 & +2.0 & +1.0 & +1.1  \\
    \bottomrule
    \end{tabular}
\end{table*}

\begin{table*}[t]
    \centering
    \caption{Results on reasoning and mathematical capability. The best results are highlighted in \textbf{bold}.}
    \label{tab:exp_mrb}
    \setlength{\tabcolsep}{1mm}
    \begin{tabular}{l c | c c  c c c c} 
    \toprule
    \textbf{Model} & \textbf{\makecell{Param \\ Size}} &  \textbf{\makecell{MathVista}}  & \textbf{\makecell{MathVerse}} & \textbf{\makecell{MathVision}} & \textbf{\makecell{WeMath}} & \textbf{\makecell{DynaMath}} & \textbf{\makecell{LogicVista}} \\
    \midrule
    \makecell[l]{GPT-4o~\cite{li2024llava}} & -  & 63.8 & 50.2 & - & - & - & -  \\
    \makecell[l]{LLaVA-OneVision~\cite{li2024llava}} & 7B  & 63.2 & 26.2 & 18.3 & 20.9 & - & 33.3  \\
    \makecell[l]{Qwen2.5-VL~\cite{bai2025qwen2}} & 3B & 62.3 & 47.6 & 21.2 & - & - & - \\
    \makecell[l]{Qwen2.5-VL~\cite{bai2025qwen2}} & 7B & 68.3 & 49.2 & 25.6 & 34.6 & 53.3 & 45.9\\
    DeepEyes \cite{zheng2025deepeyes}& 7B & 70.1 & 47.3 & 26.6 & 38.9 & 55.0 & 47.7 \\
    \midrule
    \textbf{Ours} & 7B & \textbf{72.8} & \textbf{50.6} & \textbf{28.9} & \textbf{43.5} & \textbf{57.5} & \textbf{51.0} \\
    $\Delta$ (\textit{vs} SOTA) & - & +2.7 & +0.4 & +2.3 & +4.6 & +2.5 & +3.3  \\
    \bottomrule
    \end{tabular}
\end{table*}

\begin{table*}[t]
    \centering
    \caption{Results of different decoding modes. The best results are highlighted in \textbf{bold}.}
    \label{tab:+beam_search}
    \setlength{\tabcolsep}{9mm}
    \begin{tabular}{l | c c c  } 
    \toprule
    \multirow{2}{*}{\textbf{Method}} & \multicolumn{3}{c}{\textbf{LLaVA-1.5}} \\
    % \cmidrule(lr){2-4} 
    & $\text{CHAIR}_{\rm S}$ & $\text{CHAIR}_{\rm I}$ & POPE \\
    \midrule
    DeCo & 37.8 & 11.1 & 86.7 \\
    DeCo + Beam Search & 33.0 ($\downarrow$ 4.8) & 9.7 ($\downarrow$ 1.4) & 86.7 ($\uparrow$ 0.0) \\
    HA-DPO & 38.2 & 11.0 & 82.7 \\
    HA-DPO + Beam Search & 31.7 ($\downarrow$ 6.5) & 8.9 ($\downarrow$ 2.1) & 85.2 ($\uparrow$ 2.5) \\
    GCPO & 21.5 & 5.8 & 87.2 \\
    GCPO + Beam Search & 21.1 ($\downarrow$ 0.4) & 5.5 ($\downarrow$ 0.3) & 87.5 ($\uparrow$ 0.3) \\
    \midrule
    COPO & 19.8 & 5.3 & 88.0 \\
    COPO + Beam Search & \textbf{19.7} ($\downarrow$ 0.1) & \textbf{5.3} ($\downarrow$ 0.0) & \textbf{88.1} ($\uparrow$ 0.1)\\
    \bottomrule
    \end{tabular}
\end{table*}

\begin{table*}[t]
    \centering
    \caption{Results of different GRPO variants. The best results are highlighted in \textbf{bold}.}
    \label{tab:grpo_variant}
    \setlength{\tabcolsep}{9mm}
    \begin{tabular}{l | c c c  } 
    \toprule
    \multirow{2}{*}{\textbf{Method}} & \multicolumn{3}{c}{\textbf{LLaVA-1.5}} \\
    & $\text{CHAIR}_{\rm S}$ & $\text{CHAIR}_{\rm I}$ & POPE \\
    \midrule
    DAPO \cite{liudapo} & 21.3 & 5.7 & 87.1 \\
    DAPO + COPO & 19.3 ($\downarrow$ 2.0) & 5.1 ($\downarrow$ 0.6) & 88.2 ($\uparrow$ 1.1) \\
    Dr. GRPO \cite{liu2025understanding} & 22.1 & 6.1 & 86.9 \\
    Dr. GRPO + COPO & 19.6 ($\downarrow$ 2.5) & 5.2 ($\downarrow$ 0.9) & 88.2 ($\uparrow$ 1.3) \\
    GCPO \cite{gu2025group} & 21.5 & 5.8 & 87.2 \\
    GCPO + COPO & 19.2 ($\downarrow$ 2.3) & 5.1 ($\downarrow$ 0.7) & 88.4 ($\uparrow$ 1.2) \\
    \bottomrule
    \end{tabular}
\end{table*}

\begin{table}[t]
    \centering
    \caption{Results of different implementation paths. The best results are highlighted in \textbf{bold}.}
    \label{tab:implement_path}
    \resizebox*{\linewidth}{!}{
    \begin{tabular}{l | c c c  } 
    \toprule
    \multirow{2}{*}{\textbf{Method}} & \multicolumn{3}{c}{\textbf{LLaVA-1.5}} \\
    & $\text{CHAIR}_{\rm S}$ & $\text{CHAIR}_{\rm I}$ & POPE \\
    \midrule
    COPO (Full Sequence) & 19.8 & 5.4 & 87.9 \\
    COPO (Reasoning Step) & 19.8 ($\downarrow$ 0.0) & 5.3 ($\downarrow$ 0.1) & 88.0 ($\uparrow$ 0.1) \\
    \bottomrule
    \end{tabular}}
\end{table}

\section{Additional Experiments and Full results}
\label{sec_app:experiment}

\subsection{More Evaluation Results}
To further evaluate the capability of our method, we design a series of experiments targeting distinct but complementary aspects of multi-modal model performance. High-resolution visual understanding focuses on the model's ability to accurately perceive and ground objects in ultra-high-resolution images. This evaluates fine-grained visual recognition and spatial grounding in dense scenes where hallucination is prone to occur due to the small size of objects. Grounding fidelity examines whether the model can align textual references with visual regions across diverse referential expression tasks. It tests the model's precision in resolving visual-language correspondence under different linguistic contexts. Reasoning and mathematical capability assesses the model's competence in multi-step reasoning and math-related vision-language understanding, including symbolic computation, logical inference, and numerical accuracy in visual settings. Together, these evaluations provide a broad assessment of our method's effectiveness in improving visual grounding, factual consistency, and reasoning depth under challenging multimodal scenarios. These experiments are conducted based on Qwen2.5-VL-7B-Instruct with COPO on A100 GPU clusters.

\paragraph{High-Resolution Visual Understanding.}
\label{sec_app:hrvu}
High-resolution visual understanding benchmarks, e.g. $V^*$ Bench \cite{wu2024v} and HR-Bench (4K/8K) \cite{wang2025divide}, present a significant challenge for MLLMs as they contain ultra-high-resolution images (2K-8K) where the target objects referenced in the prompts are often minuscule, sometimes occupying fewer than 200 pixels. This scenario requires both fine-grained visual recognition and precise cross-modal grounding. However, existing MLLMs frequently fail to accurately localize or reason about such small-scale targets, resulting in hallucinated outputs. \textbf{Table~\ref{tab:exp_hr}} reports the results on these benchmarks. Our approach achieves the best results among open-source MLLMs, outperforming the state-of-the-art baselines. These results reflect our method's ability to handle fine-grained visual details and complex spatial grounding in high-resolution settings.

\paragraph{Grounding Fidelity.}
\label{sec_app:gf}
We adopt several standard benchmarks to evaluate our approach for grounding. We report results on refCOCO \cite{caesar2018coco}, refCOCO+ \cite{caesar2018coco}, refCOCOg \cite{kazemzadeh2014referitgame}, and ReasonSeg \cite{lai2024lisa}. As shown in \textbf{Table \ref{tab:exp_gfhs}}, our method outperforms the base model (Qwen2.5-VL-7B) and other open-source baselines across all benchmarks. These findings suggest that our method maintains stable and accurate alignment between visual input and referential language across varied grounding benchmarks.

\paragraph{Reasoning and Mathematical Capability.}
\label{sec_app:rmc}
To further evaluate the ability in complex multimodal reasoning and mathematical understanding, we conduct experiments on several benchmark datasets, including MathVista \cite{lu2023mathvista}, MathVerse \cite{zhang2024mathverse}, MathVision \cite{wang2024measuring}, WeMath \cite{qiao2024we}, DynaMath \cite{zou2024dynamath}, and LogicVista \cite{xiao2024logicvista}. These benchmarks collectively test a wide range of capabilities, from symbolic manipulation and numerical reasoning to logical inference and visual comprehension of math-related images. As shown in \textbf{Table \ref{tab:exp_mrb}}, our method achieves strong results across all benchmarks. Notably, compared to baseline models such as Qwen2.5-VL, LLaVA-OneVision and DeepEyes, our model shows considerable gains in both general reasoning and mathematically grounded vision-language understanding. The results indicate that our method is capable of supporting structured multimodal reasoning and mathematical understanding under diverse task conditions.

\subsection{Full Results of Comparison}
\label{sec_app:comparison}
\subsubsection{Different Decoding Modes}
To further investigate the inference capability of our method, we conduct an experiment comparing DeepEyes and our method under two decoding modes: standard decoding and beam search. The evaluation is carried out on the CHAIR and POPE based on the LLaVA-1.5. We report results across $\text{CHAIR}_{\rm S}$, $\text{CHAIR}_{\rm I}$, and POPE F1 score. 

As shown in \textbf{Table~\ref{tab:+beam_search}}, DeCo, HA-DPO and GCPO demonstrates modest gains when switching from standard decoding to beam search, suggesting that more elaborate inference may help improve its performance. In contrast, our method achieves strong results without requiring additional decoding strategies, with only minor variation observed when beam search is applied. These results indicate that our method is less dependent on inference-time enhancements and may reflect stronger inference ability embedded in the model itself.

\subsubsection{Different GRPO Variants}
To further validate the generality of our causal-oriented policy optimization (COPO), we apply it on three representative GRPO variants, including DAPO \cite{liudapo}, Dr GRPO \cite{liu2025understanding}, and GCPO \cite{gu2025group}. We report results across $\text{CHAIR}_{\rm S}$, $\text{CHAIR}_{\rm I}$, and POPE F1 score based on the LLaVA-1.5. 

As reported in \textbf{Table \ref{tab:grpo_variant}}, incorporating COPO consistently improves the hallucination-related metrics on LLaVA-1.5, yielding lower $\text{CHAIR}_{\rm S}$ and $\text{CHAIR}_{\rm I}$ scores and higher POPE values.
For example, DAPO + COPO reduces $\text{CHAIR}_{\rm S}$ from 21.3 to 19.3 and $\text{CHAIR}_{\rm I}$ from 5.7 to 5.1, while GCPO + COPO achieves the best overall performance. These results demonstrate that COPO is a plug and play component for existing GRPO-based post-training schemes, which effectively constrains each decoding step to generate causally sufficient and necessary tokens, thereby mitigating hallucinations.

\subsubsection{Different Implementation Paths}
We investigate two alternative implementations of COPO: (i) full Sequence, where the causal completeness reward is uniformly applied to all output tokens; and (ii) reasoning step, where the reward is only added to reasoning tokens. Both settings are consistent with the core formulation of COPO and effectively integrate causal signals into policy optimization. We report results across $\text{CHAIR}_{\rm S}$, $\text{CHAIR}_{\rm I}$, and POPE F1 score based on the LLaVA-1.5. 

As shown in \textbf{Table \ref{tab:implement_path}}, their results are similar, indicating that either implementation can realize our COPO. Nevertheless, applying the reward only to reasoning steps yields slightly better CHAIR and POPE scores. Therefore, we adopt this implementation in our experiments.

\begin{figure}
    \centering
    \includegraphics[width=\columnwidth]{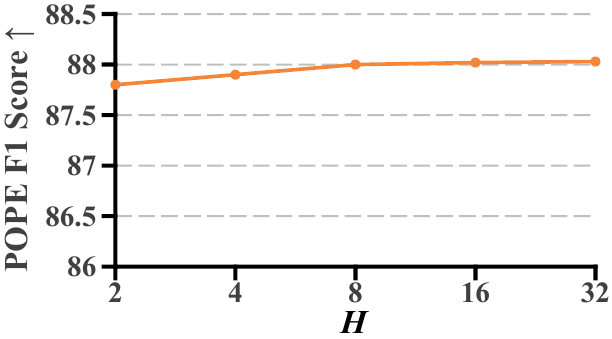}
    \caption{More parameter sensitivity.}
    \label{fig:para_h}
\end{figure}

\subsection{More Parameter Sensitivity}
We conduct experiments on the hyperparameters $H$ of Eq.\ref{eq:suff_score}. We search $H$ over $[2, 4, 8, 16, 32]$. 
\textbf{Figure~\ref{fig:para_h}} shows that the POPE F1 score exhibits a mild upward trend as $H$ increases from 2 to 8, reaching a good performance at $H=8$.  Beyond this point, further increasing $H$ yields only negligible gains while introducing additional computational overhead. This indicates that excessively large $H$ offers limited practical benefit.  Based on this trade-off between performance and efficiency, $H=8$ is selected as the final setting in our experiments.

\begin{figure}
    \centering
    \includegraphics[width=\columnwidth]{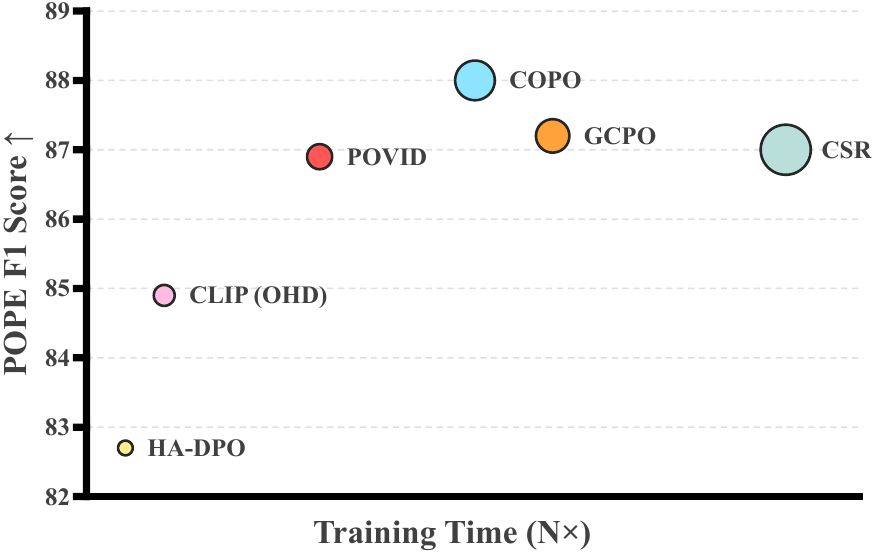}
    \caption{Trade-off performance of various methods.}
    \label{fig:tradeoff}
\end{figure}

\subsection{Trade-off Performance}
\label{sec_app:tradeoff}
We further analyze the trade-off between hallucination suppression performance and training efficiency. Following a consistent experimental setup, we compare our method with representative baselines, including HA-DPO, CLIP (OHD), POVID, CSR, and GCPO, under the same model backbone. As shown in \textbf{Figure~\ref{fig:tradeoff}}, the vertical axis indicates POPE F1 score, while the horizontal axis reflects the relative training cost (normalized training steps). Each method is marked with a distinct symbol, and our method is denoted by a blue star. The results show that our approach yields the highest F1 score with acceptable training time, indicating a balanced trade-off between performance and computational efficiency.

\end{document}